\documentclass[natbib]{article}

\usepackage{arxiv}
\usepackage[utf8]{inputenc} % allow utf-8 input
\usepackage[T1]{fontenc}    % use 8-bit T1 fonts
\usepackage{hyperref}       % hyperlinks
\usepackage{url}            % simple URL typesetting
\usepackage{booktabs}       % professional-quality tables
\usepackage{amsfonts}       % blackboard math symbols
\usepackage{nicefrac}       % compact symbols for 1/2, etc.
\usepackage{microtype}      % microtypography
\usepackage{lipsum}
\usepackage{graphicx}
\graphicspath{ {./images/} }

\title{Can RNNs learn Recursive Nested Subject-Verb Agreements?}

\author{Yair Lakretz$^*$\\Cognitive Neuroimaging Unit\\NeuroSpin center\\91191, Gif-sur-Yvette, France\\{\tt yair.lakretz@gmail.com}\\
  \And
	Théo Desbordes$^*$\\Cognitive Neuroimaging Unit\\NeuroSpin center\\91191, Gif-sur-Yvette\\Facebook AI Research\\Paris, France\\{\tt tdesbordes@fb.com}
\And
	Jean-Remi King\\Laboratoire des systèmes perceptif \\ Département d’études cognitives \\75005, École normale supérieure, \\
	PSL University, CNRS \\Facebook AI Research \\ Paris, France \\{\tt jeanremi.king@gmail.com}\\
\And
	Benoît Crabbé\\Université Paris Diderot\\Paris, France\\{\tt benoit.crabbe@gmail.com}

\And
	Maxime Oquab\\Facebook AI Research\\Paris, France\\{\tt qas@fb.com}\\

  \And
	Stanislas Dehaene\\Cognitive Neuroimaging Unit\\NeuroSpin center\\91191, Gif-sur-Yvette, \\ Collège De France \\ 75005, 11 Place Marcelin Berthelot \\ Paris, France\\{\tt stanislas.dehaene@gmail.com}
}

\usepackage{etoolbox}
\usepackage{natbib}
\usepackage{color, colortbl}
\definecolor{Gray}{gray}{0.9}
\usepackage[T1]{fontenc}
\usepackage{tikz}
\usepackage{tikz-qtree}
\usepackage{array} % for "\newcolumntype" macro
\newcommand*\OR{\ |\ }
\usepackage{amsmath}
\usepackage{subcaption}
\newcolumntype{P}[1]{>{\centering\arraybackslash}p{#1}}
\renewrobustcmd{\bfseries}{\fontseries{b}\selectfont}
\renewrobustcmd{\boldmath}{}
\newrobustcmd{\B}{\bfseries}
\usepackage{float}
\usepackage{graphicx}
\usepackage{subcaption}
\usepackage{mwe}
\usepackage{multirow}

\begin{document}

\maketitle
\def\thefootnote{*}\footnotetext{These authors contributed equally to this work.}

\begin{abstract}
One of the fundamental principles of contemporary linguistics states that language processing requires the ability to extract recursively nested tree structures. However, it remains unclear whether and how this code could be implemented in neural circuits. Recent advances in Recurrent Neural Networks (RNNs), which achieve near-human performance in some language tasks, provide a compelling model to address such questions. Here, we present a new framework to study recursive processing in RNNs, using subject-verb agreement as a probe into the representations of the neural network. We trained six distinct types of RNNs on a simplified probabilistic context-free grammar designed to independently manipulate the length of a sentence and the depth of its syntactic tree. All RNNs generalized to subject-verb dependencies longer than those seen during training. However, none systematically generalized to deeper tree structures, even those with a structural bias towards learning nested tree (i.e., stack-RNNs). In addition, our analyses revealed primacy and recency effects in the generalization patterns of LSTM-based models, showing that these models tend to perform well on the outer- and innermost parts of a center-embedded tree structure, but poorly on its middle levels. Finally, probing the internal states of the model during the processing of sentences with nested tree structures, we found a complex encoding of grammatical agreement information (e.g. grammatical number), in which all the information for multiple words nouns was carried by a single unit. Taken together, these results indicate how neural networks may extract bounded nested tree structures, without learning a systematic recursive rule.
\end{abstract}

\section{Introduction}
The syntactic organization of human languages has been proposed to follow a nested-tree organization \citep{chomsky:1957}. Nested trees necessitate a recursive operator, i.e. repeatedly applying a function over its own results. Consequently, recursive processing has been hypothesized to be at the core of the unique linguistic abilities of humans, possibly unique to humans, and is yet unidentified by human electrophysiology \citep{Hauser:etal:2002, Dehaene:etal:2015}. In recent years, Recurrent Neural Networks (RNNs) trained on large natural-language corpora have shown tremendous advances on a variety of NLP tasks, including word prediction (aka, language modeling) and sentence translation. Despite substantial differences between the human brain and artificial neural networks, the remarkable performances of modern RNNs on such tasks make them compelling objects for the study of recursive processing in neural devices akin to the human brain \citep[see, e.g., ][for early studies]{christiansen1999toward}.

The notion of recursion was developed in the study of human linguistic knowledge. If a certain construction can be generated from a given grammar by an application of a rule, then a repeated application of the same rule could generate acceptable strings of an arbitrarily complexity. In contrast, it is empirically established that human linguistic \textit{processing} is tightly limited, due to limitations such as memory capacity or attention span. This apparent inconsistency between the unbounded property of natural language and the tightly limited processing capacity of humans is commonly reconciled by drawing a distinction between human linguistic \textit{competence} and \textit{performance}. The former refers to the theoretical `ideal' knowledge of natural language, and is the object of linguistic inquiries, whereas the latter refers to the unfolding of this knowledge through parsing processes, whose operations incur a certain `cost' each. Such costs are commonly studied in psycholinguistics in behavioral experiment by measuring human accuracy and reaction times. 

In RNNs, the learned rules of the language are encoded in the network in a way that is directly related to the way the network applies them during sentence processing. Representations of abstract linguistic knowledge (construed here as `network competence'), and their unfolding in time during sentence processing (`network performance') could possibly be jointly studied in the network. \footnote{See, however, \citet{christiansen1992non}, which argue for the rejection of the competence-performance distinction in humans and RNNs.} Here, we explore the capacity of modern RNNs to learn to represent abstract rules of a recursive grammar. We study (1) the `behavioral' performance of RNNs, by evaluating model accuracy in processing nested structures sampled from the grammar, and (2) the inner representations learned by the model, by conducing an in-depth analysis into the way an RNN encodes underlying grammatical knowledge. 

Specifically, we focus on the capacity of RNN-based language models to learn artificial grammars with nested long-distance feature agreements. Feature agreement is central to our study since it allows to study recursive processing in RNNs given a linear order of words. Previous studies confirmed that RNN trained on natural data can successfully perform challenging long-range agreement between subject and verb \citep{Linzen:etal:2016, Bernardy:Lappin:2017, Gulordava:etal:2018, lakretz2019emergence}. However, several questions regarding recursive processing in RNNs remain unanswered: (1) it remains unclear whether RNNs learn to perform recursive processing over their input, akin to what argued for human language processing; (2) how do representations and mechanisms learned by RNNs affected by the statistics of the data? For example, if during training an RNN is presented with high occurrence of nested recursive structures (with, say, deeper constructions than those found in natural language), will it favor developing recursive mechanisms? (3) Can a structural bias towards learning recursive grammars (e.g., memory-augmented models) improve RNN performance on nested constructions?

To address the above questions, we introduce a setup that is simple enough to control for various aspects of the training data, which are otherwise hard to explore in a natural-data setup, while preserving a higher degree of similarity to natural data compared to previous studies on simple artificial languages. We test and compare the performance of a variety of RNN models, with and without structural bias. We found that RNNs do not genuinely capture the underlying recursive grammar and do not truly generalize to deeper structures, importantly, neither RNNs with a structural bias towards learning hierarchical data. However, RNNs do succeed in generalizing to longer subject-verb dependencies for a given depth. An analysis of the generalization patterns of the networks revealed primacy and recency effects, consistent with recent findings on the distinction between short- and long-range number units identified in RNN language models \citet{lakretz2019emergence}; Finally, we describe the dynamics of the inner states of one such long-range unit and its complex encoding of multiple grammatical numbers.
\section{Related Literature} \label{sec:relevant_lit}
Our work is closely related to the line of research started in the influential work by \citet{Linzen:etal:2016}. Since then, grammatical long-distance agreement has become a standard way to probe the syntactic capabilities of RNN language models \citep{Bernardy:Lappin:2017, Gulordava:etal:2018, lakretz2019emergence}, and was extended to other related phenomena \citep{jumelet2018language, futrell2018rnns, marvin2018targeted, wilcox2019syntactic} and to other types of neural models \citep[e.g., ][]{goldberg2019assessing}.

Studying syntax processing with natural language incurs difficulties arising from various correlations between syntactic and semantic information. In contrast, artificial languages provide a framework for studying the purely syntactic abilities of RNNs while carefully controlling for various parameters of the data. Starting with the classic work on Simple Recurrent Networks (SRN) in \citet{Elman:1991}, studies explored the ability of RNNs to learn simple grammars. Initially, moving from regular to context-free grammars (CFGs), difficulties have been reported in learning such languages with SRNs \citep[][e.g.,]{boden1999learning, boden2000context}. Later, Long-Short Term Memory (LSTM) networks have been found to better capture simple context-free and context-sensitive counter languages, such as $a^nb^n$ and $a^nb^nc^n$, showing robust generalization to longer sequences (\citealp{Gers:Schmidhuber:2001, rodriguez2001simple}; \citealp[see also, ][]{weiss2018practical, suzgun2018evaluating}). 

More recently, LSTM networks were evaluated on their capacity to capture Dyck languages.\footnote{A Dyck-$n$ language consists of strings with balanced pairs of $n$ different types of brackets. For example, $"[ \{ \} ] \{ \{ [ ] \} \}"$ is an admissible sequence of a Dyck-2 language. 
Dyck languages are both simple and expressive enough to represent all CFGs \citep{chomsky1959algebraic} (in particular the Dyck-2 language suffices \citep{suzgun2019lstm}).} 
Dyck-1 was found to be well captured by LSTM networks \citep{bernardy2018can}, in particular, \citet{suzgun2019lstm} showed that a single-layer LSTM with a single hidden unit suffices to recognize the language. 
More recently, \citet{suzgun2019memory} have shown that memory-augmented RNN models can capture generalized Dyck languages, including the Dyck-2 language, with close to perfect out-of-sample performance. 
Other related work studied the generalization patterns of seq2seq models in a synthetic language setup \citep{Lake:Baroni:2017}, pointing out failures of all models to generalize in a systematic way to unseen data \citep[see also, ][]{Hupkes:etal:2017, hupkes2019compositionality}

% Indeed, this language can be captured with a one-counter machine. 
% However, capturing the Dyck-2 language remained a demanding task for all models in this study \citep[see also, ][]{sennhauser2018evaluating}, and also for seq2seq attention-based models \citep{yu2019learning}. 

\section{Models} \label{sec: models}
We contrasted two families of models - with and without a structural bias:

\subsection{Standard RNNs}\label{ssec:rnns}
We explored three standard RNN units: (1) Simple Recurrent Networks (SRNs), introduced by \citet{Elman:1991}, (2) Short-Term Memory (LSTM) models \citep{Hochreiter:Schmidhuber:1997}, which were previously shown to perform well on number-agreement tasks when trained on a natural-language corpus, and (3) Gated Recurrent Unit (GRU) \citep{cho2014learning}. In contrast to LSTM units, GRUs merge the two state variables $h$ and $C$ into a single one, which makes the handling of interference materials less explicit (but possible). The coupling of the input and output gate was shown to prevent the network from developing counting mechanisms to recognize simple context-free grammars \citep{weiss2018practical}. 

\subsection{RNNs with a structural bias}\label{ssec:structured_rnns}
\paragraph{Ordered-Neurons LSTMs (ON-LSTMs)} Based on the intuition that larger constituents contain information that changes more slowly across the sentence, \citet{shen2018ordered} suggested a variant of 
LSTMs, called Ordered-Neurons LSTMs, which imposes a hierarchical bias on the cell-updating mechanism. Given the hierarchical nature of our data, we expected ON-LSTMs to perform well on the number-agreements tasks. In our experiments, we used the publicly available code.\footnote{https://github.com/yikangshen/Ordered-Neurons. We made a single modification - we replaced the loss criterion in the code by a standard cross-entropy loss, which is more appropriate for our simple setup.}

\paragraph{Stack-RNNs and Stack-LSTMs} Stack-RNNs are memory augmented RNNs, which learn to perform a sequence of 'soft' pop and push actions on a stack-like array \citep{Joulin:Mikolov:2015}. Stack-LSTMs are a variant of Stack-RNNs, with a modified update of the hidden state: $h_t = LSTM(x_t, \tilde{h}_{t - 1})$ \citep[see, e.g., ][]{suzgun2019memory}. Given the high performance of these models on counter languages and Dyck-2, we tested the performance of these models on our agreement tasks, using the code provided by the authors\footnote{https://github.com/suzgunmirac/marnns. To have a tighter comparison with the other models, we made the following two modifications: (1) since we trained all models with a language modelling objective (section \ref{sec:setup}), we added a softmax layer to the output layer, and (2) we added an embedding layer between the input and first layer of the model, and between the last layer and the softmax.}.
\section{Experimental Setup} \label{sec:setup}
We trained all our models with a language-modelling objective and tested them on a number prediction task \citep{Linzen:etal:2016, Gulordava:etal:2018, futrell2018rnns}. To explore the effect of the statistics and properties of the data on the resulting dynamics of the model, we generated training datasets with different characteristics. We did so by sampling from a pCFG with two parameters that separately control the mean tree depth and dependency lengths in the data. The models were then tested on well-controlled number-agreement tasks, which allow us to probe subject-verb number agreement in controlled and increasingly challenging way.
%\footnote{All datasets and codes will be made available upon publication.}

\begin{figure}
\begin{center}
\begin{tikzpicture}[sibling distance=0.5pt]
\tikzset{edge from parent/.append style={very thick}}
\Tree [.S [.NP [.A $a_3$ ] [.A $a_2$ ] [.N $\textbf{n}_5[s]$ ] ] [.S [.NP [.A $a_1$ ] [.N $\textbf{n}_1[p]$ ] ] [.VP [.A $a_2$ ] [.A $a_4$ ] [.V $\textbf{v}_3[p]$ ] ] ] [.VP [.A $a_4$ ] [.V $\textbf{v}_2[s]$ ] ] ]
\end{tikzpicture}
\end{center}
\caption{An example for a sentence sampled from the pCFG in (1).} %(\ref{grammar}).}
\label{fig:tree}
\end{figure}
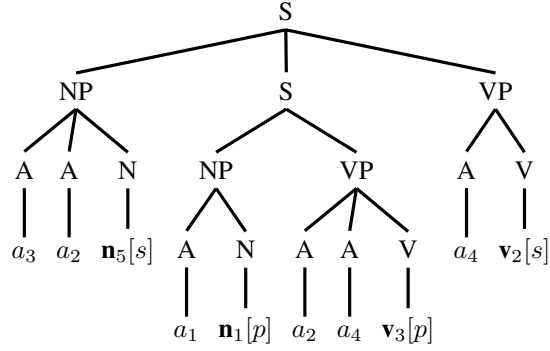

\paragraph{The Grammar}
The training sets were sampled from the following center-embedding probabilistic CFG with feature agreement:

\begin{align} %\label{grammar}
    \begin{split}
    \textbf{S}&\to \textbf{NP}\mkern6mu \textbf{VP}\mkern6mu \{1-p1\} \OR \textbf{NP}\mkern6mu \textbf{S}\mkern6mu \textbf{VP}\mkern6mu \{p1\}\\
    \textbf{NP} &\to \textbf{N}\mkern6mu \{1-p2\} \OR A\mkern6mu \textbf{NP}\mkern6mu \{p2\}\\
    \textbf{VP} &\to \textbf{V}\mkern6mu \{1-p2\} \OR A\mkern6mu \textbf{VP}\mkern6mu \{p2\}\\
    \textbf{N} &\to \textbf{n}_1 \mkern6mu \{0.2\} \OR \textbf{n}_2 \mkern6mu \{0.2\} \OR ... \OR \textbf{n}_5\mkern6mu \{0.2\}\\
    \textbf{V} &\to \textbf{v}_1 \mkern6mu \{0.2\} \OR \textbf{v}_2 \mkern6mu \{0.2\} \OR ... \OR \textbf{v}_5\mkern6mu \{0.2\}\\
    A &\to a_1 \mkern6mu \{0.2\} \OR a_2 \mkern6mu \{0.2\} \OR ... \OR a_5\mkern6mu \{0.2\}
    \end{split}
\end{align}

where, $S$, $NP$ and $VP$ represent start, noun-phrase and verb-phrase-like non-terminals. In curly brackets, $p_1$ and $p_2$ are the generation probabilities of a center-embedded clause and adjective-like preceding non-terminal, respectively. $n_i$ and $v_i$ are noun- and verb-like terminal tokens that carry grammatical number, and $a_i$ are adjective-like terminal tokens that do not carry number. Finally, $N$, $V$ and $A$ are their corresponding part-of-speech (noun, verb and adjective). We highlight in bold feature agreement between the left-hand and right-hand side of each production; in this simplified CFG, there is no agreement for adjectives, and therefore $A$ and $a_i$ variables are not in bold. The addition of A tokens to the grammar is crucial since it allows to increase sentence length (by changing $p_2$) without changing depth (which is only affected by $p_1$). Figure \ref{fig:tree} shows an example of a training sample and its corresponding tree. Note that the numbering $i$ of the various tokens $n_i$, $v_i$ and $a_i$ is not related to the depth of the token.

\begin{figure*}[ht!]
    \centering
    \includegraphics[height=7cm, width=11cm]{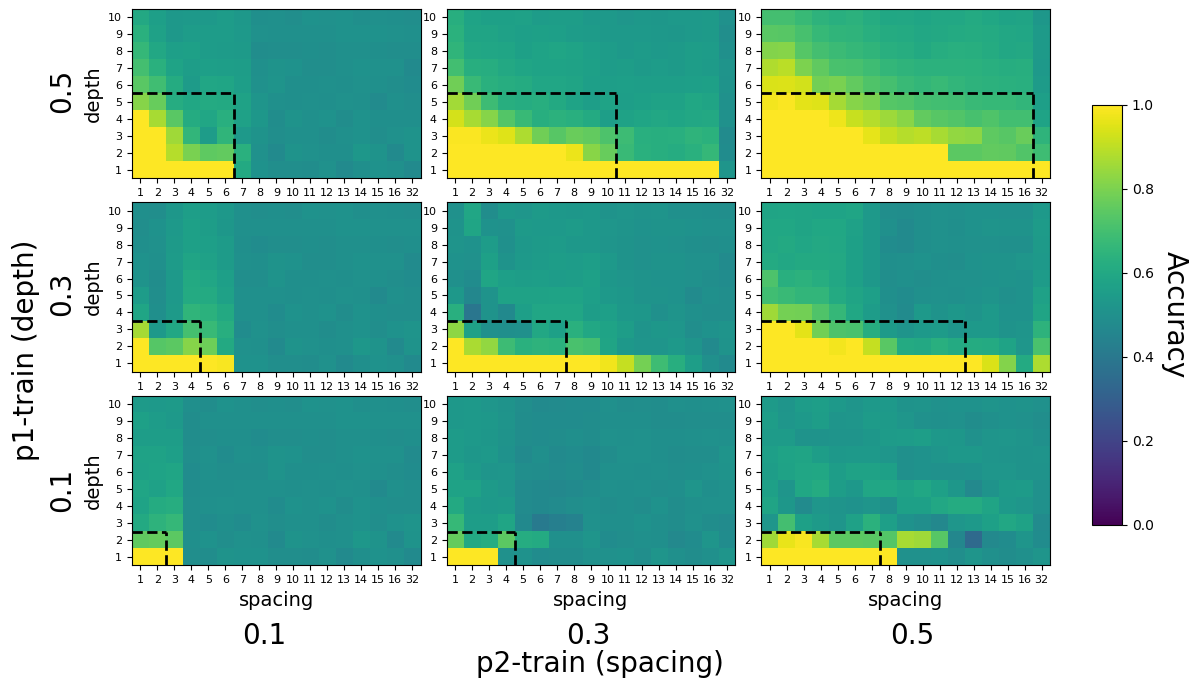}
    \caption{Average accuracy across all verbs and sentences in the number-agreement tasks. Each matrix corresponds to one of the nine LSTM models $M_{p_1, p_2}$. Each pixel corresponds to accuracy on a specific NA-task of a given depth and spacing $(d, s)$. The maximal accuracy is one, which corresponds to the case in which the model predicted the right grammatical number of all verb-like tokens in all 1K sentences. Dashed horizontal and vertical black lines represent the maximal training depth and spacing.}
    \label{fig:accuracy_all_verbs}
\end{figure*}

\paragraph{Training datasets} We generated different training datasets by sampling from the grammar with various combinations of generation probabilities: $p_1, p_2 \in \{0.1, 0.3, 0.5\}$. This resulted in nine training sets, and we set the number of tokens in each to 1M. To control for the maximal length and depth of sentences presented to the models, we truncated the tail of these distributions by clipping the maximal values to its $95^{th}$ percentile. In what follows, we refer to each training set by $D_{p_1, p_2}^{train}$.

\paragraph{Validation Datasets} Validation datasets were generated in the same way with 100K each. We tuned the following hyperparameters: number of layers, hidden units per layer, embedding size and dropout. In what follows, we refer to the optimal model from each training set as $M_{p_1, p_2}$.

\paragraph{Test Datasets} Similarly, for model comparison, for each probability pair, we generated a test set with 200K tokens.

\paragraph{Number-Agreement Tasks (NA-tasks)} We tested the performance of the models on fixed-length NA-tasks of increasing difficulty. Each NA-task is composed of $d$ nested dependencies and a fixed spacing $s$, which is the number of successive adjective-like tokens between any two number-carrying tokens (noun or verb-like tokens). For example, for $d=3$, $s=2$, one of the samples is: 
'a2 a1 \textbf{n3[sg]} a5 a3 \textbf{n1[pl]} a2 a2 \textbf{v5[pl]} a4 a1 \textbf{v[sg]} a2 a5', where number-carrying tokens are in bold. Note that noun- and verb-like tokens at each depth agree on their number. Note also that each sentence is padded with an affix of $s=2$ successive adjective-like tokens on both sides. For each pair of $d \in (1, 2, ..., 10)$ and $s \in (1, 2, 3, ..., 16, 32)$, we generated 1K sentences, resulting in a total of 170 test datasets. 

\paragraph{Model Evaluation}
For hyperparamter tuning and model comparison, we used perplexity to evaluate the performance of the models on the validation set. For model evaluation, we tested model performance on the NA-tasks: similarly to \citet{Linzen:etal:2016}, for each task, we presented the sentences sequentially, one word at a time, and compared the prediction of the model to the correct grammatical number one step before each verb in the sentence. 
%Since the lexical-production distributions of the pCFG are uniform, instead of comparing singular vs. plural forms of a specific verb token, we compared the predictions after separately averaging over all singular and plural forms. 
Accuracy in a given NA-task was measured as the proportion of sentences for which the model assigned a higher likelihood to the correct verb form.

\section{Experiments} \label{sec:experiments}
\begin{figure*}[ht!]    
    \centering
    \includegraphics[height=8cm, width=9cm]{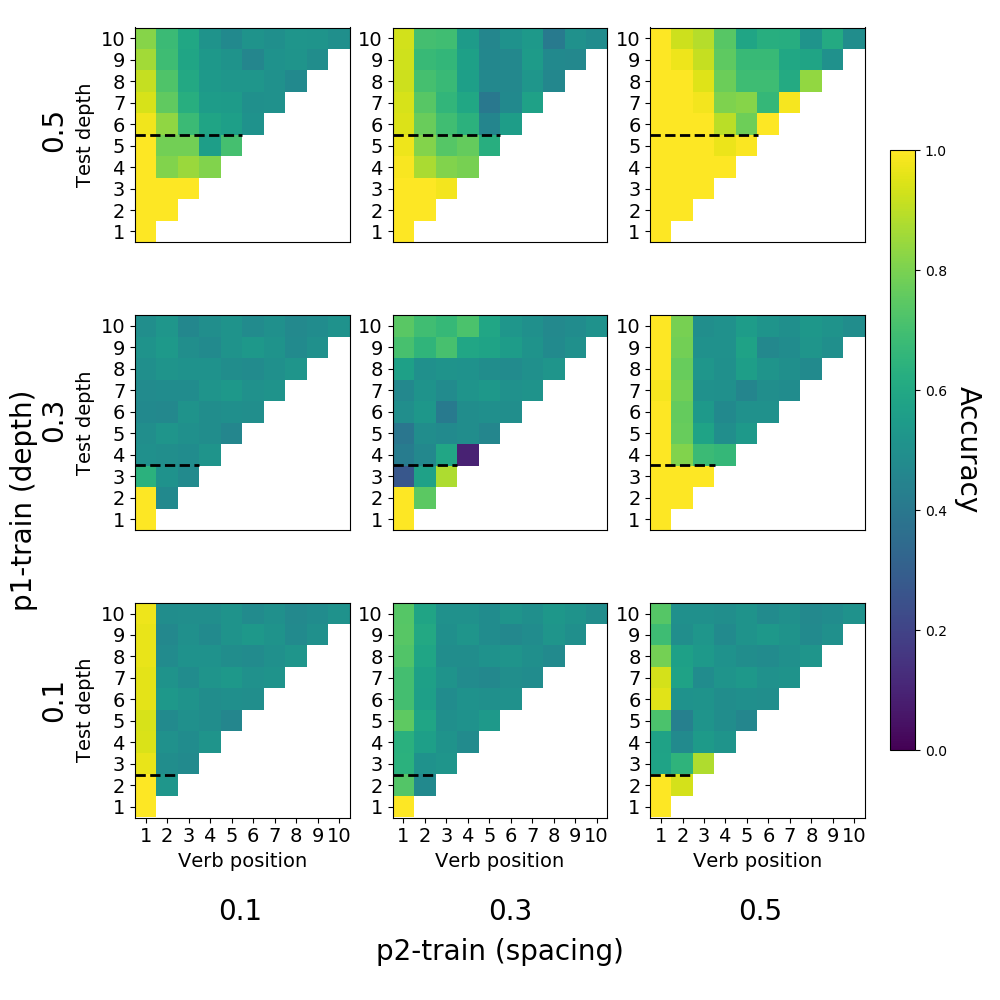}
    \caption{Accuracy per verb for the nine models on NA-tasks with spacing equals to two ($s=2$). Verb position is with respect to counting from the inner- to outermost verb in the center-embedding structure. Dashed horizontal lines correspond to the maximal training depth.}
    \label{fig:accuracy_per_verbs}
\end{figure*}

We start by studying the LSTM model and then extend our analyses to all other models (section \ref{ssec:model_comparison}). The experiments in this section provide the following observations:
\begin{itemize}
    \itemsep0em 
    \item LSTMs do not generalize to deeper structures, but do generalize to longer dependencies (\ref{ssec:generalization}). 
    \item LSTMs show recency and primacy effects in their generalization patterns (\ref{ssec:perverb}).  
    \item A small set of units explains most of the dynamics of the LSTM when performing the agreement task (\ref{ssec:mechanism}).
    \item All other RNN models fail to generalize to deeper structures, but do succeed to generalize to longer sequences. Recency and primacy effects were found also in models with a structural bias (\ref{ssec:model_comparison}). 
\end{itemize}

\subsection{Generalization of Standard LSTMs}\label{ssec:generalization}
We trained the models as described in section \ref{sec:setup}, which resulted in nine models $M_{p_1, p_2}$ for the nine training datasets. We then evaluated the models on each of the NA-tasks. Figure \ref{fig:accuracy_all_verbs} describes the resulting accuracies for all LSTM models. We first note that models that were trained on deeper datasets (larger $p_1$) achieve better performance on sentences with greater depth. Similarly, models trained on longer datasets (larger $p_2$) better perform with respect to spacing. Second, we note that the performance of all models is relatively low for a set of $d$ and $s$ values that remain below their maximal training values. This highlights the importance of a careful evaluation of model performance that probes the structural representations captured by the model, rather than performance evaluation based on perplexity alone \citep[see e.g., ][]{suzgun2018evaluating, goldberg2019assessing}. Third, model performance shows an interaction between depth and spacing - for larger values of $d$, the accuracy of the models decreases faster with spacing. This is expected due to increased interference among the multiple grammatical numbers that need to be carried over time for nested long-range dependencies. Fourth, for $d=1$, most models generalize beyond the max-training spacing, in particular, $M_{0.5, 0.5}$ generalize from max-training spacing of 16 to 32. Finally, we note that the generalization of all models to deeper structures is qualitatively poor, with above-chance performance only in some cases. As we will show next, also in these cases, the models do not truly generalize to deeper structure, but rather perform well on only the same number of verbs encountered during training. 

\begin{table*}[ht!]
\centering
\begin{tabular}{|P{0.9cm}||P{1cm}|P{1cm}|P{1cm}|P{1cm}|P{1cm}|P{1cm}|P{1cm}|P{1cm}|P{1cm}|P{1cm}|}
% &&&&&&&&&& \\
\hline
 & \multicolumn{10}{c|}{\B Verb position} \\
\cline{1-11}
 & \multicolumn{2}{c|}{1}& \multicolumn{2}{c|}{2}& \multicolumn{2}{c|}{3}& \multicolumn{2}{c|}{4}& \multicolumn{2}{c|}{5}\\
\hline
\B Depth & S & P & S & P & S & P& S & P& S & P \\
\hline
1 & - & - & \cellcolor{Gray} & \cellcolor{Gray} & \cellcolor{Gray} & \cellcolor{Gray} & \cellcolor{Gray} & \cellcolor{Gray} & \cellcolor{Gray} & \cellcolor{Gray} \\
\hline
2 & - & - & 23 & - & \cellcolor{Gray} &  \cellcolor{Gray} & \cellcolor{Gray} & \cellcolor{Gray} & \cellcolor{Gray} & \cellcolor{Gray}\\
\hline
3 & - & - & 23 &  - & 23 & - & \cellcolor{Gray} & \cellcolor{Gray} & \cellcolor{Gray} & \cellcolor{Gray}\\
\hline
4 & - & - & - & - & 23 & - & 4,23 & - & \cellcolor{Gray} & \cellcolor{Gray} \\
\hline
5 & - & - & - & - & 23 & 12 & 4,23,26 & 5,6,7 & 23 & 5,7 \\
\hline

\end{tabular}
\caption{Results from the ablation study: numbers in cells represent units whose ablation brings the performance of the network to close or below chance level ($<0.55$).} \label{table:ablation}
\end{table*}

\begin{figure*}[ht!]
    \centering
    \begin{subfigure}{\textwidth}
            \centering
            \includegraphics[height=1cm, width=0.98\linewidth]{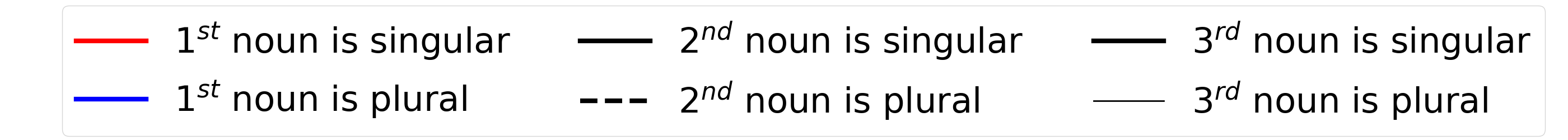}
    \end{subfigure}
    
    \begin{subfigure}{0.49\textwidth}
            \centering
            \includegraphics[height=2.3cm, width=\linewidth]{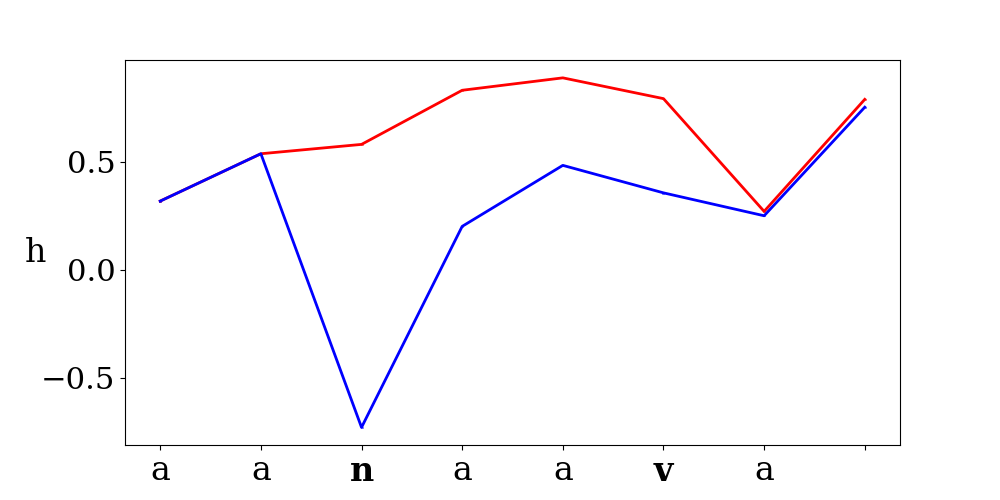}
            \subcaption{A single dependency ($d=1, s=2$).} 
    \label{fig:unit23_d1_s2}
    \end{subfigure}
    \begin{subfigure}{0.49\textwidth}
            \centering
            \includegraphics[height=2.3cm, width=\linewidth]{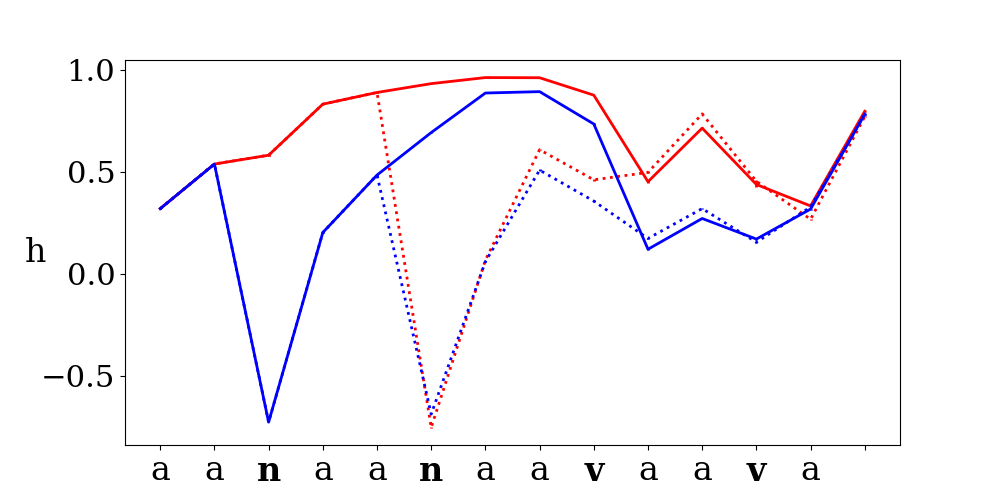}
            \subcaption{Two dependencies ($d=2, s=2$).}
    \label{fig:unit23_d2_s2}
    \end{subfigure}
    \begin{subfigure}{0.49\textwidth}
            \centering
            \includegraphics[height=2.3cm, width=\linewidth]{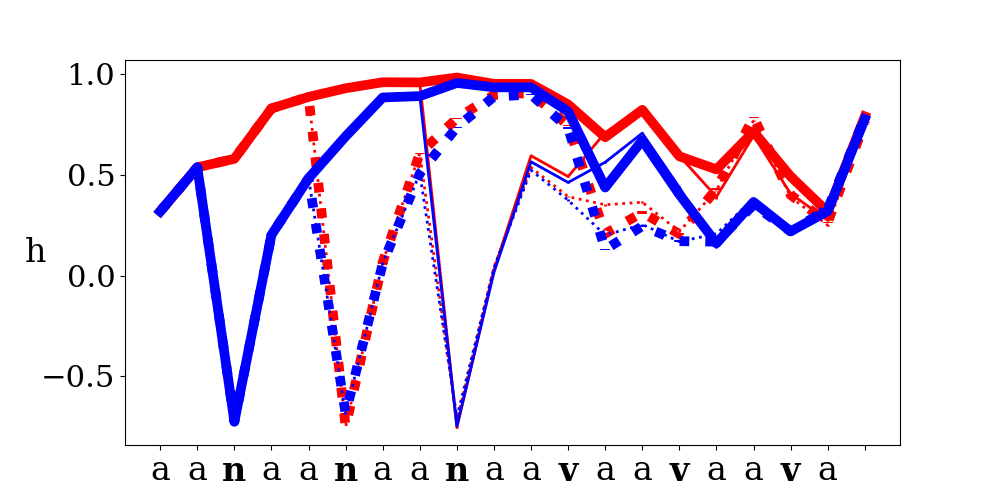}
            \subcaption{Three dependencies ($d=3, s=2$).}
    \label{fig:unit23_d3_s2}
    \end{subfigure}
    \begin{subfigure}{0.49\textwidth}
        \centering
        \includegraphics[height=2.3cm, width=\linewidth]{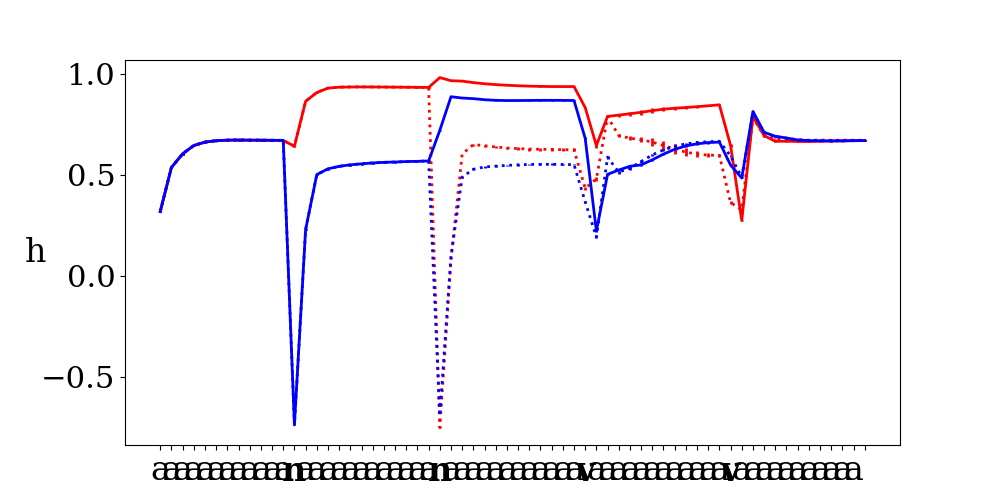}
        \subcaption{Two dependency with long spacing ($d=2, s=12$)} 
        \label{fig:unit23_d2_s12}
    \end{subfigure}
    % \bigskip
    \begin{subfigure}{0.93\textwidth}
            \centering
            \includegraphics[height=3cm, width=\linewidth]{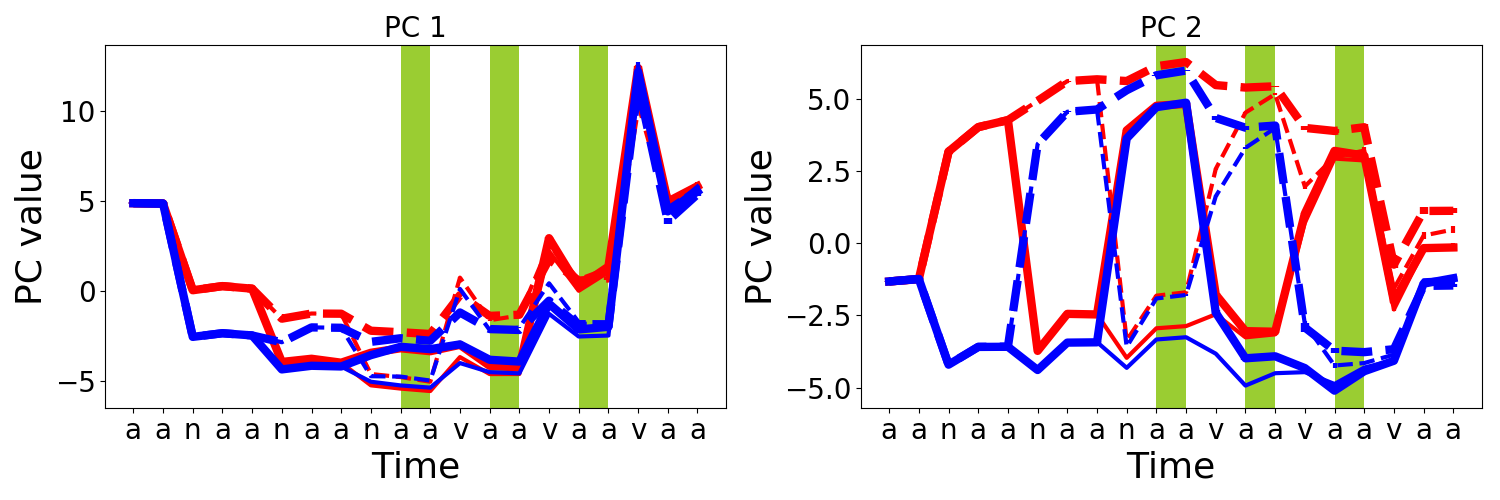}
            \subcaption{PCA analysis of the hidden states. Left: first PC, Right: second PC.} 
    \label{fig:pca_hidden}
    \end{subfigure}
    \caption{Internal dynamics of unit 23 from $M_{0.5, 0.5}$. Lines represent average values across all sentences in the NA task. We use line color, style and weight to encode the number of each noun: red and blue colors represent that the first noun is either singular or plural, respectively; continuous and dashed lines represent that the second noun is either singular or plural, respectively; thick and thin lines represent that the third nouns is either singular or plural, respectively. In (e), 1$^{st}$, 2$^{nd}$ and 3$^{rd}$ shaded areas highlight the time steps before verb prediction, in which the trajectories for the first grammatical number (line widths), second (line styles) and third (colors) are separated. }
\end{figure*}

\subsection{Primacy and Recency effects in LSTMs}\label{ssec:perverb}
To better characterize the regularization patterns of the models, we analyzed the performance of the models on each verb in the sentence. Figure \ref{fig:accuracy_per_verbs} shows the accuracy of the models on the NA-tasks on each verb separately, for a fixed spacing $s=2$. We first note that in most cases the models generalize to much deeper sentences when considering the first verb only, or the first few verbs (i.e. the innermost ones in the center-embedded structure). This is compatible with the emergence of two types of number units in the network: short and long-range units \citep{lakretz2019emergence}. In this study, it was found that many short-range units encode and carry grammatical number for several time steps if there is no interference from an opposite grammatical number in the input, whereas long-range units are few but can robustly carry the grammatical number across possible interference. We refer to this as the recency effect, which suggests that the model can encode the last encountered noun(s). Next, we note that $M_(0.5, 0.5)$ performs well on the last verb for $d>5$, although its performance is poor on the middle verbs. We refer to this effect as 'primacy' effect, which suggests that the model encodes the first noun in the sentence for long-range dependencies. To test the robustness of this effect, we analyzed other models, including a model that was trained on higher statistics and other models from the grid-search with similar hyperparameters. All showed the same effect. In sum, LSTM models show primacy and recency effects, but no satisfactory generalization to deep center-embedded structures.

\subsection{Single-unit Dynamics during Sentence Processing} \label{ssec:mechanism}

To identify units in the network that encode grammatical number for long-range dependencies, we ran an ablation study on $M_{0.5, 0.5}$. We successively ablated each individual unit and evaluated the ablated model on NA-tasks with $d \in (1, 2, 3, 4, 5)$ and $s=2$, as in the previous section, analyzing the performance on each verb separately. To further understand the encoding of singular vs. plural in each noun position, we split the sentences in the NA-task with respect to whether the noun was plural or singular before evaluating the ablation effect. Table \ref{table:ablation} summarizes the results. We first note that none of the units caused a dramatic reduction in performance on the inner-most verb (verb 1) when ablated. This recency effect is again consistent with the existence of multiple short-range units that encode the last grammatical number in a distributed and redundant way. Second, ablating unit 23 consistently reduced the performance of the network across several depths and verb positions when the noun was singular. As for plural, we found that only for $d=5$ were there units that could dramatically impair the performance of the network, although we found that for smaller depths the performance was significantly impaired by the same units as for $d=5$, but not as much as lowering to around chance level.

The ablation study suggests that unit 23 has a key role in performing the NA task. To understand the encoding strategy of this unit, we analyzed its dynamics during the processing of sentences from the NA-tasks. Figure \ref{fig:unit23_d1_s2}-\ref{fig:unit23_d3_s2} shows the hidden state dynamics during the processing of sentences from the NA-task with $s=2$ for depths one to three. For $d=1$, unit 23 separately encodes the grammatical number of the noun (red vs. blue) throughout the subject-verb dependency. For $d=2$, the unit separately encodes the number of the first noun (red vs. blue) one step before it is required to be predicted, before the outer verb, and separately encodes the number of the second noun (continuous vs. dashed lines) one step before the inner verb should be predicted. Similarly, for $d=3$, the two line colors are distinguishable before the outer verb, the two lines styles before the middle one, and the two line weights (thick vs. thin) are distinguishable before the innermost one. These dynamics are consistent with the results of the ablation study, which shows that unit 23 has a dramatic effect on network performance both on the outer and the middle verb in $d=3$. Finally, we explore the dynamics of unit 23 when it fails to perform the NA-task. Figure \ref{fig:unit23_d2_s12} shows its dynamics on the NA-task with $(d, s)=(2,12)$. In contrast to $s=2$ above, larger spacing prevents the network from encoding the number of the first nouns for long-range dependencies (blue and red lines are not well separated previous to one step before the outer verb). In sum, this analysis shows that complex encoding of three successive nouns can be carried out by a single unit in the network. 

To visualize the dynamics of also the entire network during the processing of deep nested dependencies, we calculated the first two principal components (PCs) of the state space of the network at each time step. Figure \ref{fig:pca_hidden} shows the results for the PCA of the hidden states of the network during the NA-task with $(d, s)=(3,2)$. We found a similar encoding dynamics to that used by unit 23, with both PCs showing separated trajectories for the different grammatical numbers before each verb prediction, showing that the network follows a similar encoding scheme. Finally, we note that both the PCA and dynamics of unit 23 show a 'fractal-like' encoding, in which information about all previous features is simultaneously stored at different scales. To exemplify this, note that after the third noun (first shaded area in PC2), information about the previous two nouns is preserved at smaller scales: the number of the first noun is encoded with higher values for singular compared to plural (red lines are consistently above the blue), and similarly for the number of the second noun (dashed lines are above the continuous). Such a fractal, multiscale activity pattern may provide a general solution to the neural implementation of a memory stack and the associated push and pop operations.

\subsection{Model Comparison} \label{ssec:model_comparison}
\begin{figure*}[ht!]    
    \centering
    \includegraphics[height=5.5cm, width=\linewidth]{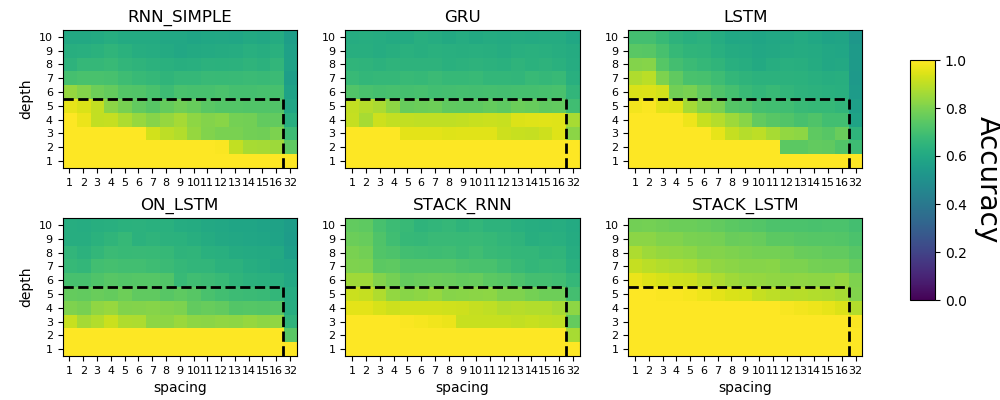}
    \caption{Average accuracy across all verbs and sentences in each NA-task, for all models when trained on $D_{0.5, 0.5}^{train}$.}
    \label{fig:all_models}
\end{figure*}

Next, we evaluated all other types of RNNs in the same way we did for the LSTM. We first found that, overall, memory-augmented models achieved the best perplexity - all other models had comparable performance. We next evaluated all models on the NA-tasks: Figure \ref{fig:all_models} shows the average accuracies on all NA-tasks for all types of models trained on $D_{0.5, 0.5}^{train}$. We first note that stack-LSTMs show better overall performance also on the NA-tasks, achieving perfect performance on almost all depths and spacing values within the max limits of the training set, and showing generalization to longer dependencies for all depths $d<4$. However, we note that none of the models genuinely generalized to deeper structures, as can be also observed in the generalization patterns of the models (see SM). Interestingly, we found that all models with a structural bias (ON-LSTMs, Stack-RNNs and Stack-LSTMs) showed recency and primacy effects in their generalization patterns, similarly to LSTMs. This suggests that this property is common to many RNN models.

\section{Summary and Discussion}
We presented a new framework for studying the ability of neural language models to learn recursively nested long-range dependencies. 
We made use of two independent control parameters of a center-embedding pCFG to decouple generalization patterns to deeper structures from those to longer dependencies. 
Our framework thus diverges from previous work \citep{christiansen1999toward, christiansen2001finite, suzgun2019memory} by studying long-range dependencies in a more naturalistic setup, with a richer vocabulary, syntactic features defined over multiple tokens, models that are trained with a language-modeling objective and with a focus on subject-verb agreement. 

Comparing performance of various RNN models, we found that none of the models truly captured the underlying recursive regularity.
In particular, structured models - ON-LSTMs, stack-RNNs and stack-LSTMs - were found to be comparable, or only superficially better, than standard RNNs and did not genuinely generalize to deeper structures. While stack-RNNs were recently shown to achieve excellent performance on the Dyck-2 language \citep{suzgun2019memory}, we found that they fail to generalize in a more demanding setup, even when an extensive optimization search in hyperparameter space is performed. In contrast, for small depths already encountered during training, mostly for $d=1$, all models generalized well to longer sentences, showing that they learned to well capture long-range dependencies up to a given depth, in accordance to previous studies on natural data.

Our ablation study revealed that while the innermost dependency is redundantly encoded, performance on outer dependencies can be dramatically reduced by ablations of single units. Indeed, we found that a single unit in the network can robustly encode all grammatical numbers of a sentence of depth three. This provides further evidence for the emergence of sparse mechanisms for long-range dependencies in LSTM language models \citep{lakretz2019emergence}.

Several of our models (LSTM, ON-LSTM, stack-RNN and stack-LSTM) showed recency and primacy effects in their generalization patterns, with stronger recency effects compared to primacy. 
The recency effect corresponds to the ability of the models to robustly encode the grammatical feature of the most recently encountered nouns, and is consistent with the existence of short-range number units in LSTM-LMs. 
The primacy effect corresponds to the ability of the model to robustly encode the grammatical feature of the first noun in the sentence, and is consistent with the finding of long-range number units in LSTM-LMs \citep{lakretz2019emergence}. 
Performance on the middle nouns was found to be relatively low, suggesting that these models tend to better encode the outer and innermost parts of the syntactic tree. 
Intriguingly, human performance shows similar patterns in various cognitive tasks. For example, recency and primacy effects were reported in free-recall, with typically stronger recency effects compared to primacy ones \citep[e.g., ][]{murdock1962serial}. More closely related to our setup, in psycholinguistics, it was found that subjects tend to judge sentences with doubly nested relative clause structures as grammatical correct even though these sentences were missing a verb - this phenomenon is known as 'structural forgetting' \citep{gibson1999memory}. Importantly, structural forgetting occurs only with sentences in which the \textit{middle} verb is omitted. The similarity between the generalization patterns of many of our RNN models and those reported in humans is therefore striking - we intend to further explore this in future work. 

\section*{Acknowledgments}
We would like to thank Marco Baroni, Dieuwke Hupkes, German Kruszewski and Christophe Pallier for helpful feedback and comments on the work. This work was supported by the Bettencourt-Schueller Foundation and an ERC grant, "NeuroSyntax" project, to SD; and the Fyssen Foundation, grant ANR-17-EURE-0017 to JRK through his role with PSL University.

\bibliographystyle{plainnat}
\bibliography{yair.bib}

\end{document}

\maketitle

\section{Training Details}
\label{sec:training}

We explored with the following hyperparameters when training all the models:
\begin{itemize}
    \setlength\itemsep{0.05em}
    \item number of layers: 1, 2, 4
    \item number of hidden units per: 4, 8, 16, 32
    \item embedding size: 4, 8
    \item dropout: 0.1, 0.3
    \item chunk size (ON-LSTM only): 1, 4
\end{itemize}

% What's more, we tested values 1 and 4 for the chunk size parameter, specific to the ON\_LSTM architecture. This parameter controls the size of the groups of units that share the same master gates and hence update together. 

Back propagation through-time length and batch size were set to 32. We used the ADAM optimizer with learning rate $1e^{-3} $ and trained each model for 20 epochs (except for stack-RNNs and stack-LSTMs for which 3 epochs were enough for convergence). The optimal model was determined based on validation-set perplexity. 

%%%%%%%%%%%%%%%%%%%%%%%%%%%%%%%%%%%%%%%%%%%%%%%%%%%%%%%
%%%%%%%%%%%%%%%%    TRAIN DATA STATS   %%%%%%%%%%%%%%%%
%%%%%%%%%%%%%%%%%%%%%%%%%%%%%%%%%%%%%%%%%%%%%%%%%%%%%%%

\begin{figure*}[h!]
    \centering
    \includegraphics[height=16cm, width=\textwidth]{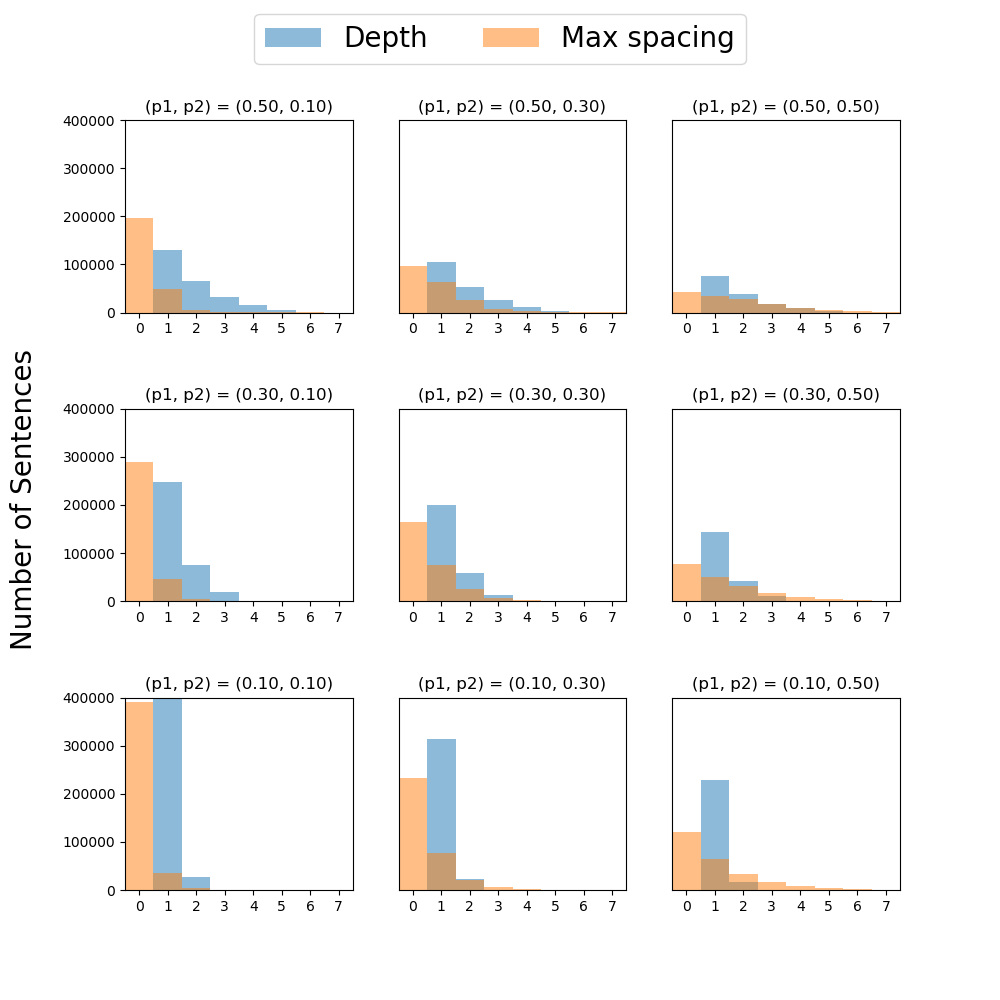}
    \caption{}
    \label{fig:srn_gen}
    \caption{Training datasets statistics. Each diagram shows the distribution of depth and maximum spacing corresponding to one of the (p1, p2), 1000000 tokens training datasets. Note that the total number of sentences per dataset is smaller for the higher probabilities because sentences are longer.} 
\end{figure*}

%%%%%%%%%%%%%%%%%%%%%%%%%%%%%%%%%%%%%%%%%%%%%%%%%%%%%%%
%%%%%%%%%%%%%%%%    PERPLEXITY     %%%%%%%%%%%%%%%%%%%%
%%%%%%%%%%%%%%%%%%%%%%%%%%%%%%%%%%%%%%%%%%%%%%%%%%%%%%%

\begin{figure*}[h!]
    \centering
    \includegraphics[height=16cm, width=\textwidth]{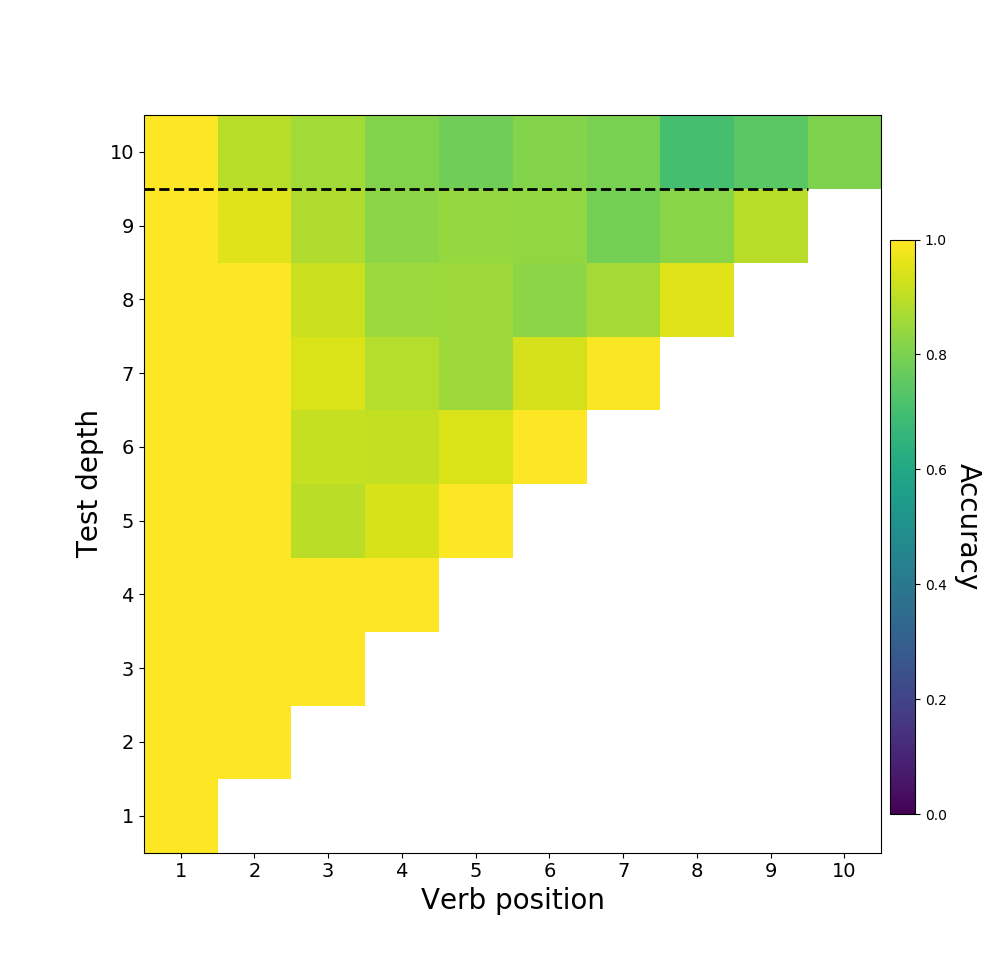}
    \caption{}
    \label{fig:ppl}
    \caption{\textbf{LSTMs:} Accuracy for $M_{0.7, 0.7}$ on the number-agreement tasks with spacing equals to two ($s=2$). Each pixel corresponds to accuracy on a specific verb in the sentence when tested on a NA-task with a given depth $d$.} 
\end{figure*}

%%%%%%%%%%%%%%%%%%%%%%%%%%%%%%%%%%%%%%%%%%%%%%%%%%%%%%%
%%%%%%%%%%%%%%%%    CELL DYNAMICS 23   %%%%%%%%%%%%%%%%
%%%%%%%%%%%%%%%%%%%%%%%%%%%%%%%%%%%%%%%%%%%%%%%%%%%%%%%

\begin{figure*}[ht!]
    \centering
    \begin{subfigure}{\textwidth}
            \centering
            \includegraphics[width=0.98\linewidth]{figures/legend_d3_s2.png}
    \end{subfigure}
    % \bigskip
    \begin{subfigure}{0.49\textwidth}
            \centering
            \includegraphics[height=3cm, width=\linewidth]{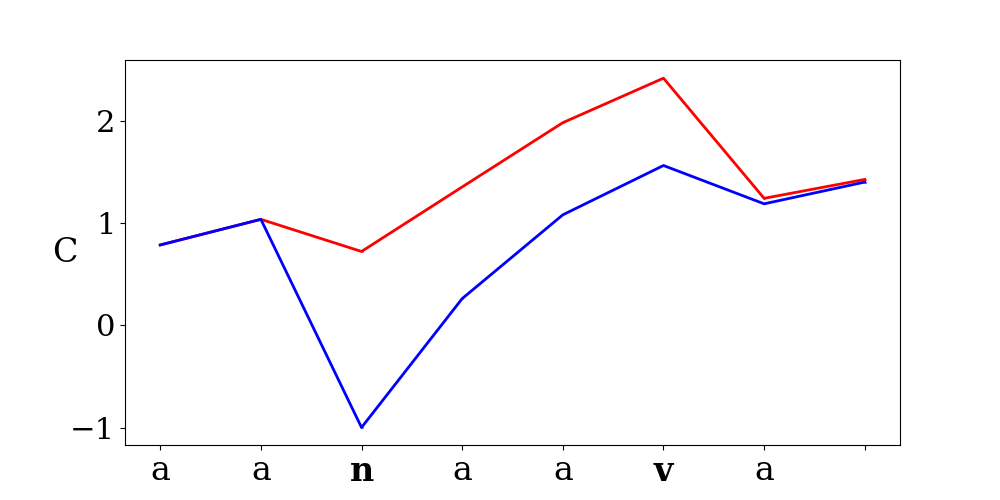}
            \subcaption{A single dependency ($d=1, s=2$).} 
    \label{fig:23_d1_s2}
    \end{subfigure}
    \begin{subfigure}{0.49\textwidth}
            \centering
            \includegraphics[height=3cm, width=\linewidth]{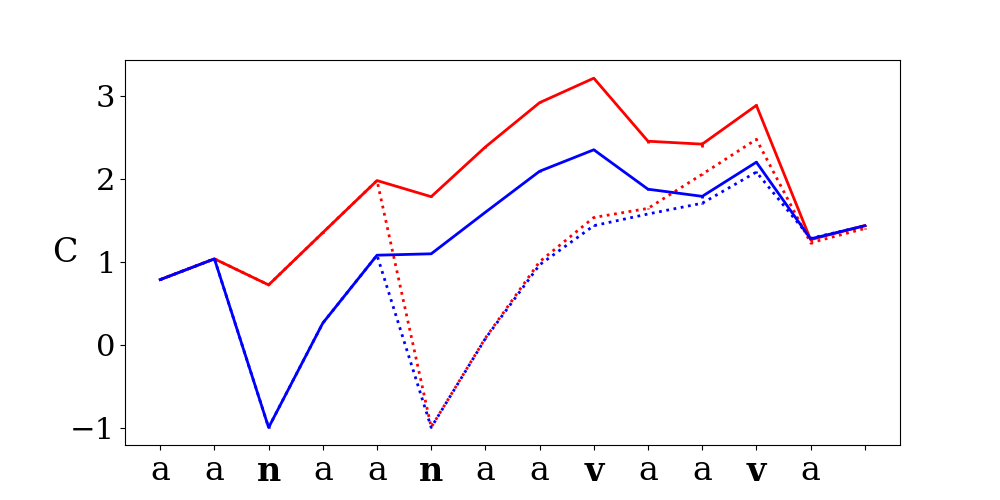}
            \subcaption{Two dependencies ($d=2, s=2$).}
    \label{fig:23_d2_s2}
    \end{subfigure}
    \begin{subfigure}{0.49\textwidth}
            \centering
            \includegraphics[height=3cm, width=\linewidth]{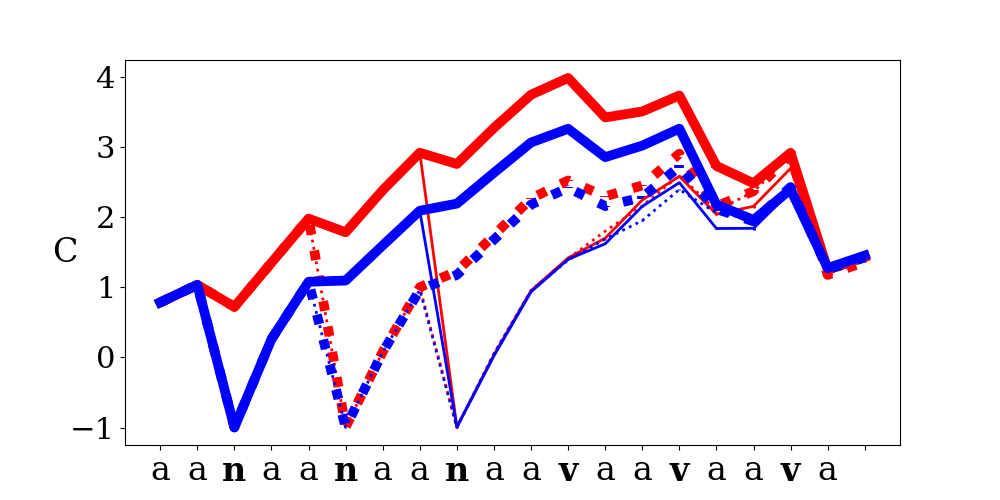}
            \subcaption{Three dependencies ($d=3, s=2$).}
    \label{fig:23_d3_s2}
    \end{subfigure}
    \begin{subfigure}{0.49\textwidth}
        \centering
        \includegraphics[height=3cm, width=\linewidth]{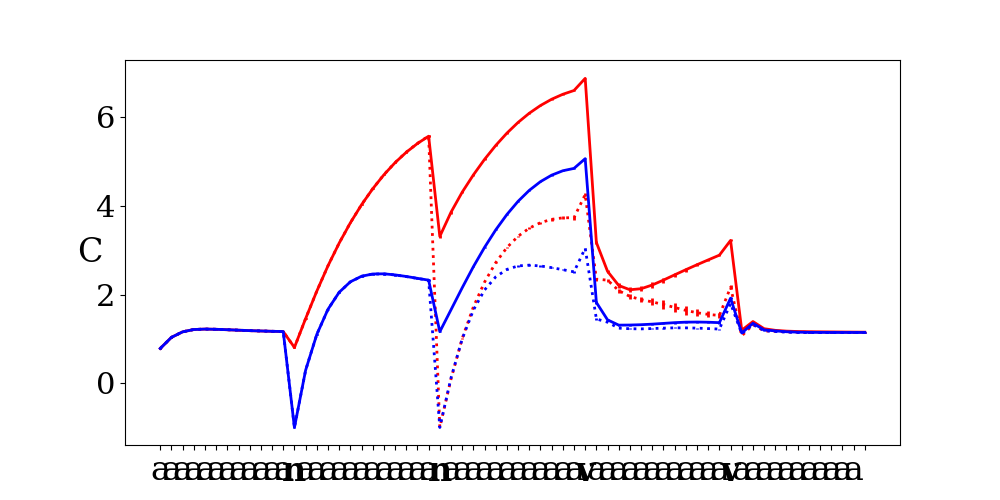}
        \subcaption{Two dependency with long spacing ($d=2, s=12$)} 
        \label{fig:23_d2_s16}
    \end{subfigure}
    \caption{Cell dynamics of unit 23 from $M_{0.5, 0.5}$. Lines represent average values across all sentences in the NA task. We use line color, style and weight to encode the number of each noun: red and blue colors represent that the first noun is either singular or plural, respectively; continuous and dashed lines represent that the second noun is either singular or plural, respectively; thick and thin lines represent that the third nouns is either singular or plural, respectively.}
\end{figure*}

%%%%%%%%%%%%%%%%%%%%%%%%%%%%%%%%%%%%%%%%%%%%%%%%%%%%%%%
%%%%%%%%%%%%%%%%    PCA       %%%%%%%%%%%%%%%%%%%%%%
%%%%%%%%%%%%%%%%%%%%%%%%%%%%%%%%%%%%%%%%%%%%%%%%%%%%%%%

\begin{figure*}[ht!]
    \centering
    \begin{subfigure}{\textwidth}
            \centering
            \includegraphics[width=0.98\linewidth]{figures/legend_d3_s2.png}
    \end{subfigure}
    % \bigskip
    
    \begin{subfigure}{0.98\textwidth}
            \centering
            \includegraphics[height=3cm, width=\linewidth]{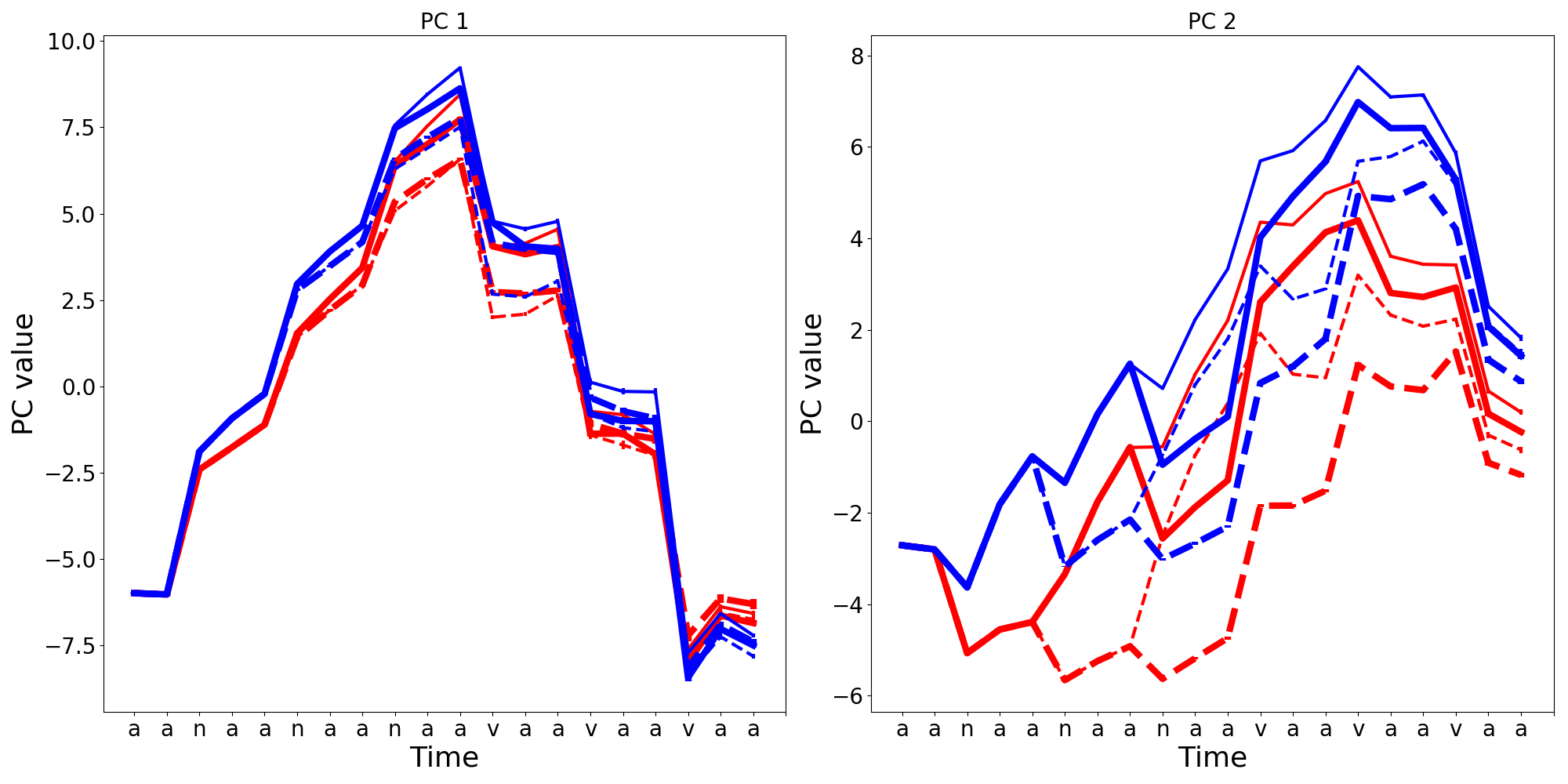}
        \subcaption{Cell states} 
    \label{fig:pca_cell}
    \end{subfigure}
    
    \caption{PCA analysis of cell states for the NA-task: $d=3$, $s=2$. Note that the first PC shows a counter-like dynamics.}
\end{figure*}

%%%%%%%%%%%%%%%%%%%%%%%%%%%%%%%%%%%%%%%%%%%%%%%%%%%%%%%
%%%%%%%%%%%%%%%%    PERPLEXITY     %%%%%%%%%%%%%%%%%%%%
%%%%%%%%%%%%%%%%%%%%%%%%%%%%%%%%%%%%%%%%%%%%%%%%%%%%%%%

\begin{figure*}[h!]
    \centering
    \includegraphics[height=16cm, width=\textwidth]{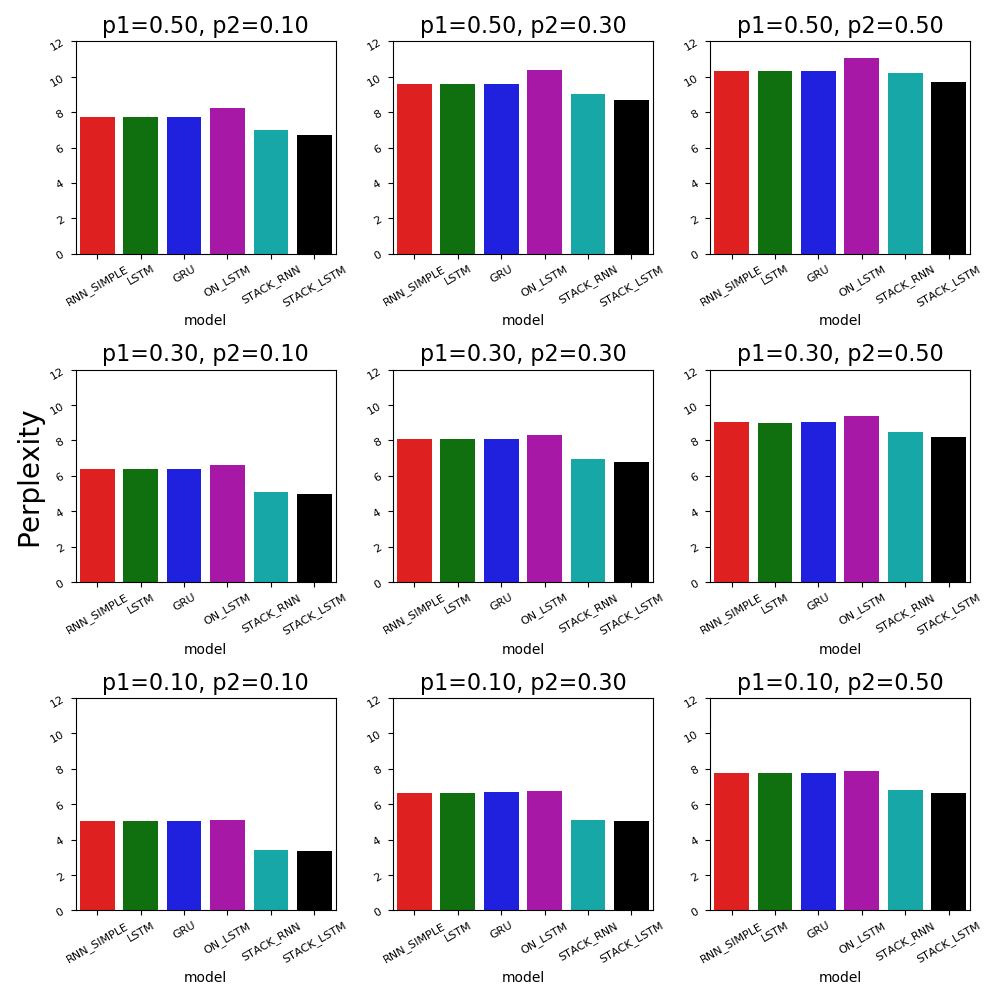}
    \caption{}
    \label{fig:ppl}
    \caption{Test perplexity for all models. Each diagram corresponds to a $(p1, p2)$-training dataset and show the perplexity achieved by each model on the corresponding test dataset.} 
\end{figure*}

%%%%%%%%%%%%%%%%%%%%%%%%%%%%%%%%%%%%%%%%%%%%%%%%%%%%%%%
%%%%%%%%%%%%%%%%    ACCURACIES     %%%%%%%%%%%%%%%%%%%%
%%%%%%%%%%%%%%%%%%%%%%%%%%%%%%%%%%%%%%%%%%%%%%%%%%%%%%%

\begin{figure*}[h!]
    \centering
    \includegraphics[height=10cm, width=\textwidth]{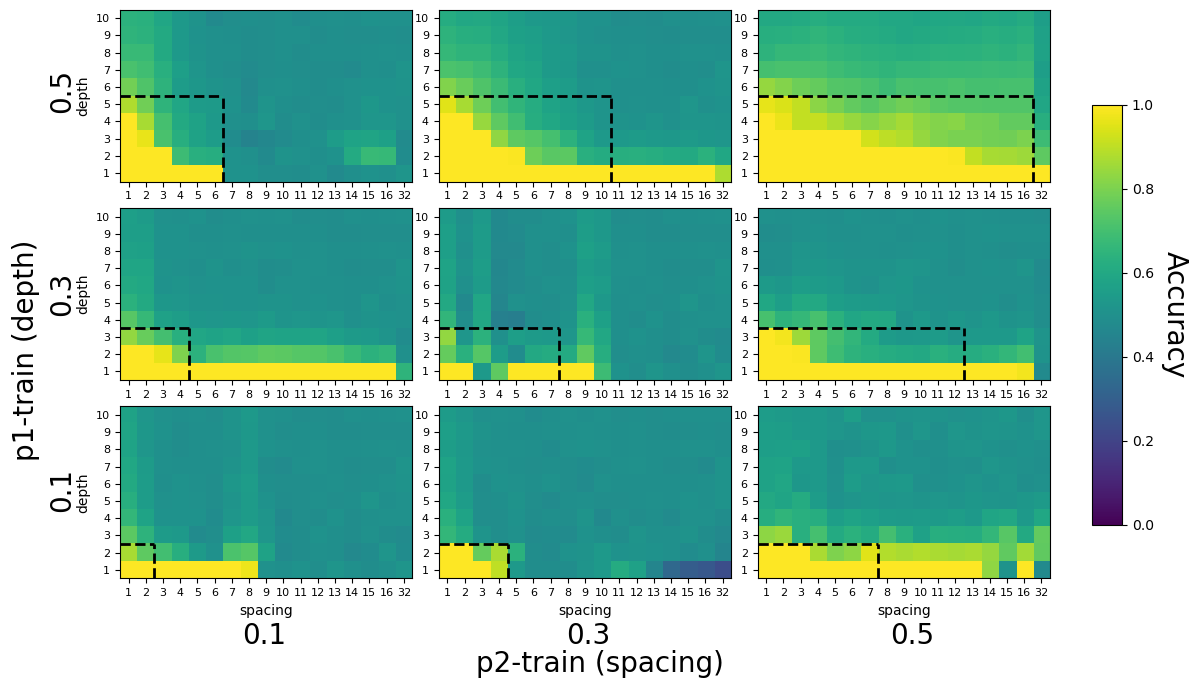}
    \caption{\textbf{SRNs:} Average accuracy across all verb on the number-agreement tasks, for each of the nine models. Each matrix corresponds to a model trained on one of the $(p1, p2)$-training datasets. Each pixel in a matrix corresponds to accuracy on a specific NA-task with a given $(d, s)$. Dashed horizontal and vertical black lines correspond to the maximal depth and spacing observed during training, respectively.}
    \label{fig:srn_gen}
\end{figure*}

\begin{figure*}[h!]    
    \centering
    \includegraphics[height=12cm, width=\textwidth]{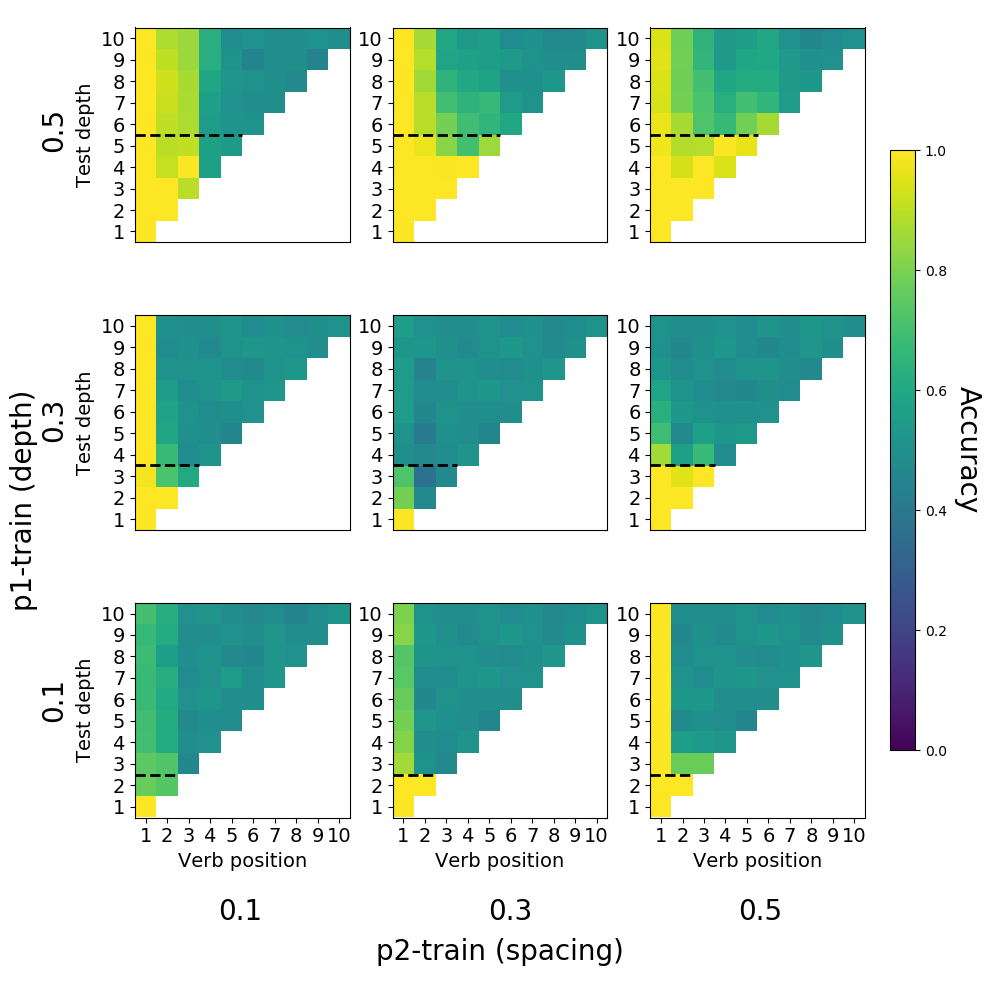}
    \caption{\textbf{SRNs:} Accuracy for the nine models on number-agreement tasks with spacing equals to two ($s=2$). Each matrix corresponds to a model trained on one of the $(p1, p2)$-training datasets. Each pixel in a matrix corresponds to accuracy on a specific verb in the sentence when tested on a NA-task with a given depth $d$. Dashed horizontal lines correspond to the maximal depth observed during training.}
    \label{fig:}
\end{figure*}

\begin{figure*}[h!]
    \centering
    \includegraphics[height=10cm, width=\textwidth]{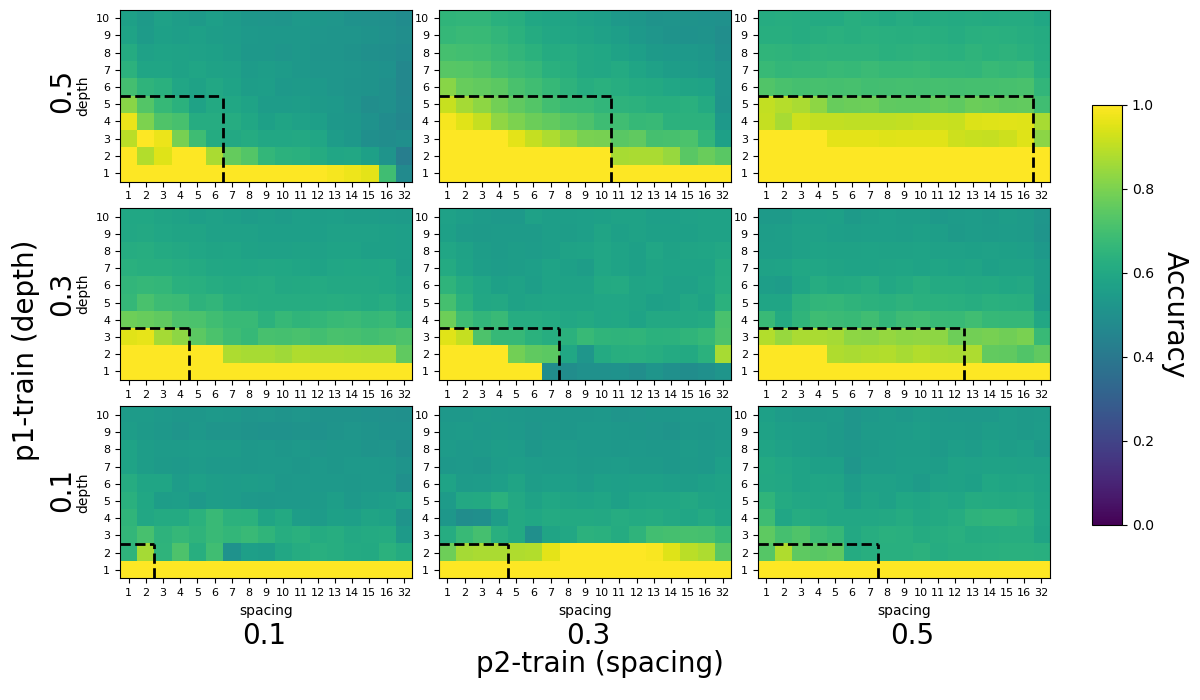}
    \caption{\textbf{GRUs:} Average accuracy across all verb on the number-agreement tasks, for each of the nine models. Each matrix corresponds to a model trained on one of the $(p1, p2)$-training datasets. Each pixel in a matrix corresponds to accuracy on a specific NA-task with a given $(d, s)$. Dashed horizontal and vertical black lines correspond to the maximal depth and spacing observed during training, respectively.}
    \label{fig:}
\end{figure*}

\begin{figure*}[h!]    
    \centering
    \includegraphics[height=12cm, width=\textwidth]{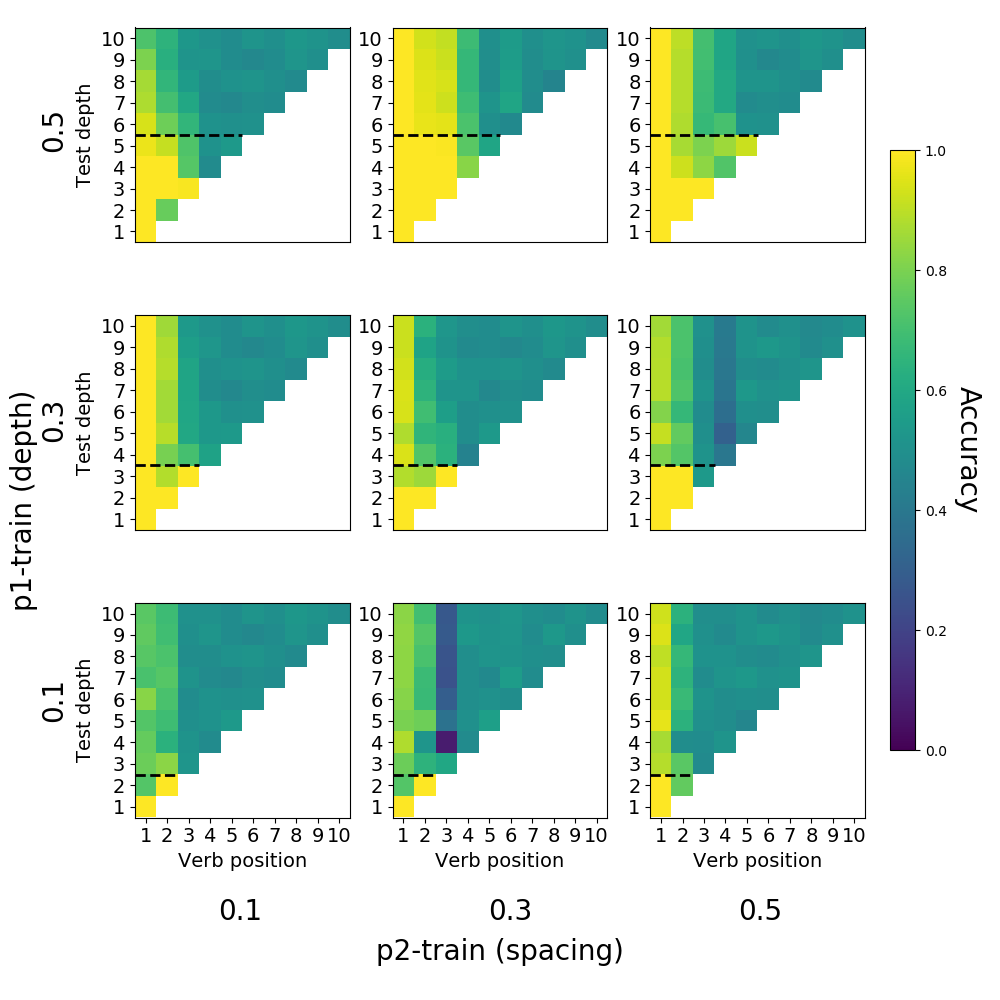}
    \caption{\textbf{GRUs:} Accuracy for the nine models on number-agreement tasks with spacing equals to two ($s=2$). Each matrix corresponds to a model trained on one of the $(p1, p2)$-training datasets. Each pixel in a matrix corresponds to accuracy on a specific verb in the sentence when tested on a NA-task with a given depth $d$. Dashed horizontal lines correspond to the maximal depth observed during training.}
    \label{fig:}
\end{figure*}

\begin{figure*}[h!]
    \centering
    \includegraphics[height=10cm, width=\textwidth]{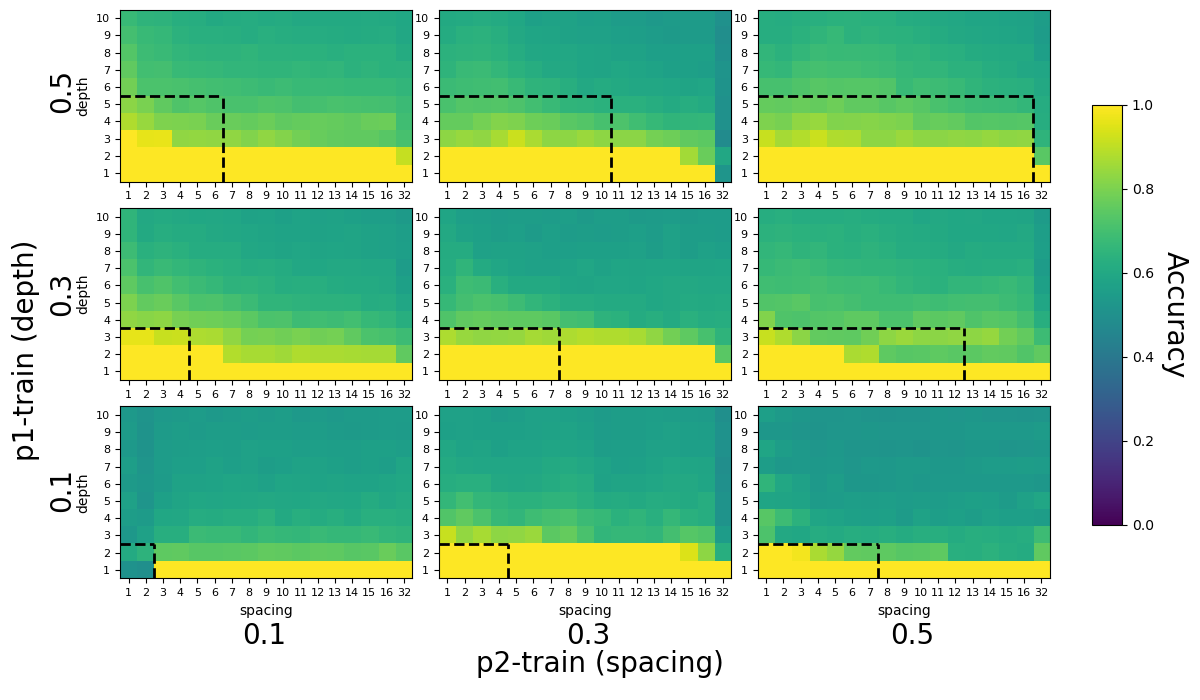}
    \caption{\textbf{ON-LSTMs:} Average accuracy across all verb on the number-agreement tasks, for each of the nine models. Each matrix corresponds to a model trained on one of the $(p1, p2)$-training datasets. Each pixel in a matrix corresponds to accuracy on a specific NA-task with a given $(d, s)$. Dashed horizontal and vertical black lines correspond to the maximal depth and spacing observed during training, respectively.}
    \label{fig:}
\end{figure*}

\begin{figure*}[h!]    
    \centering
    \includegraphics[height=12cm, width=\textwidth]{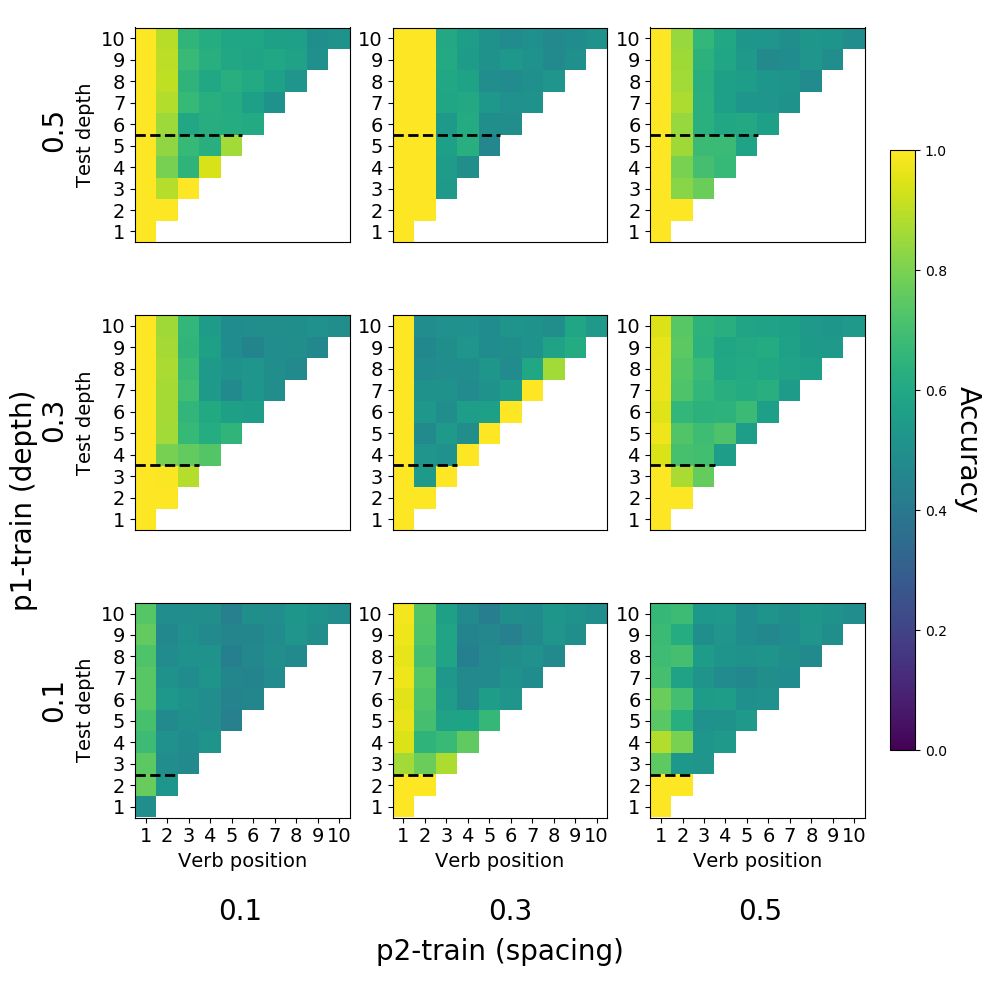}
    \caption{\textbf{ON-LSTMs:} Accuracy for the nine models on number-agreement tasks with spacing equals to two ($s=2$). Each matrix corresponds to a model trained on one of the $(p1, p2)$-training datasets. Each pixel in a matrix corresponds to accuracy on a specific verb in the sentence when tested on a NA-task with a given depth $d$. Dashed horizontal lines correspond to the maximal depth observed during training.}
    \label{fig:}
\end{figure*}

\begin{figure*}[h!]
    \centering
    \includegraphics[height=10cm, width=\textwidth]{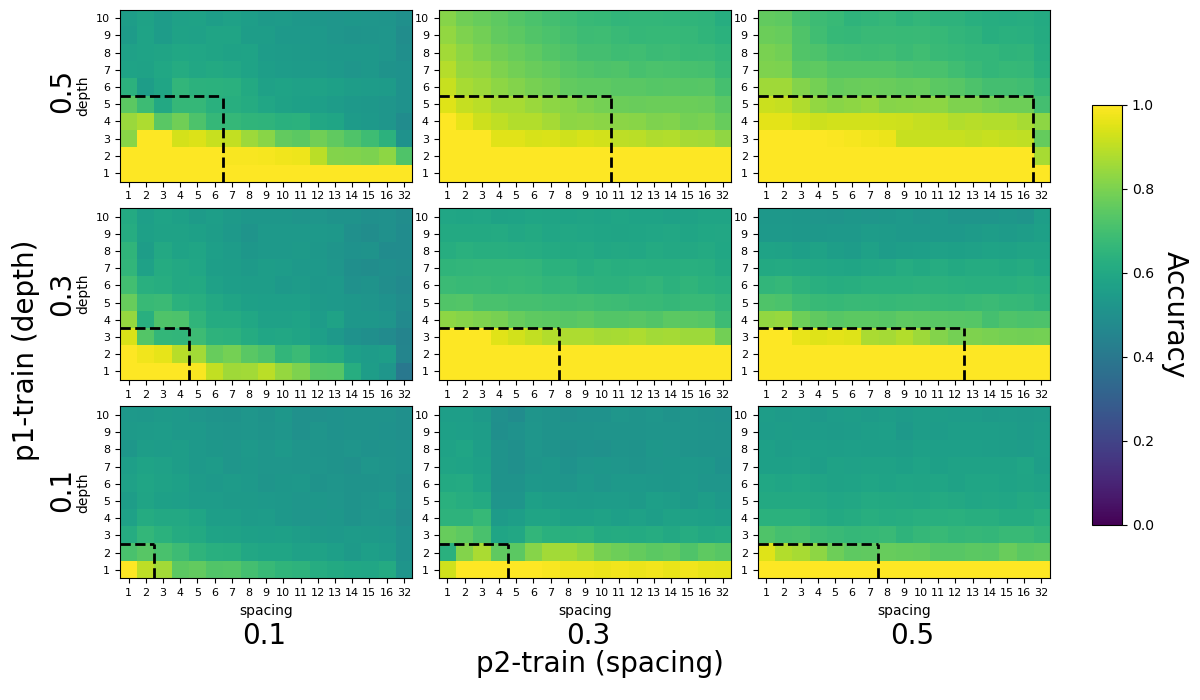}
    \caption{\textbf{Stack-RNNs: }Average accuracy across all verb on the number-agreement tasks, for each of the nine models. Each matrix corresponds to a model trained on one of the $(p1, p2)$-training datasets. Each pixel in a matrix corresponds to accuracy on a specific NA-task with a given $(d, s)$. Dashed horizontal and vertical black lines correspond to the maximal depth and spacing observed during training, respectively.}
    \label{fig:}
\end{figure*}

\begin{figure*}[h!]    
    \centering
    \includegraphics[height=12cm, width=\textwidth]{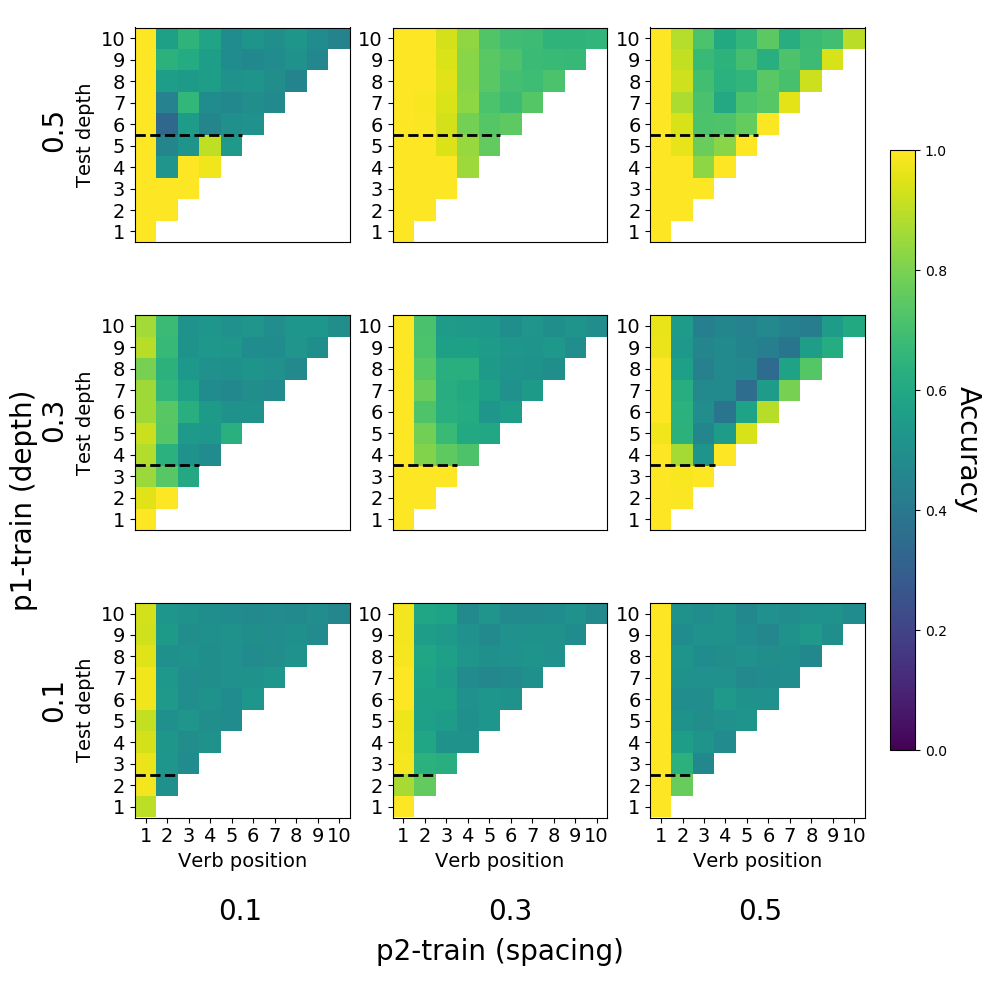}
    \caption{\textbf{Stack-RNNs: }Accuracy for the nine models on number-agreement tasks with spacing equals to two ($s=2$). Each matrix corresponds to a model trained on one of the $(p1, p2)$-training datasets. Each pixel in a matrix corresponds to accuracy on a specific verb in the sentence when tested on a NA-task with a given depth $d$. Dashed horizontal lines correspond to the maximal depth observed during training.}
    \label{fig:}
\end{figure*}

\begin{figure*}[h!]
    \centering
    \includegraphics[height=10cm, width=\textwidth]{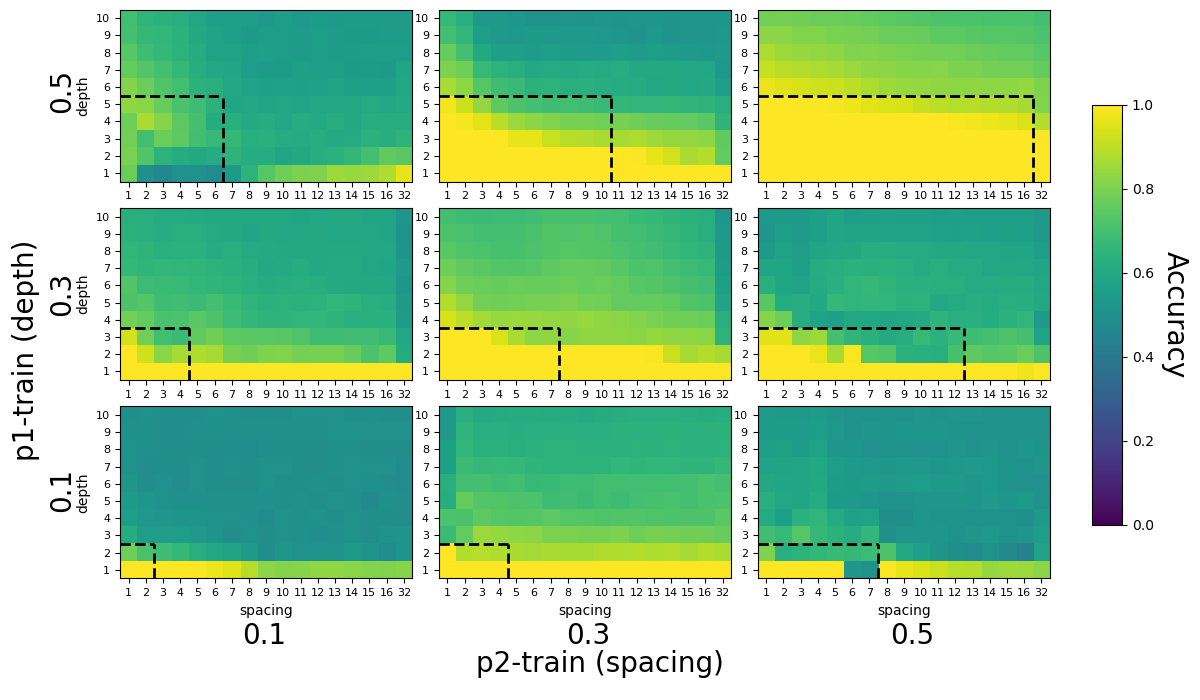}
    \caption{\textbf{Stack-LSTMs: }Average accuracy across all verb on the number-agreement tasks, for each of the nine models. Each matrix corresponds to a model trained on one of the $(p1, p2)$-training datasets. Each pixel in a matrix corresponds to accuracy on a specific NA-task with a given $(d, s)$. Dashed horizontal and vertical black lines correspond to the maximal depth and spacing observed during training, respectively.}
    \label{fig:}
\end{figure*}

\begin{figure*}[h!]    
    \centering
    \includegraphics[height=12cm, width=\textwidth]{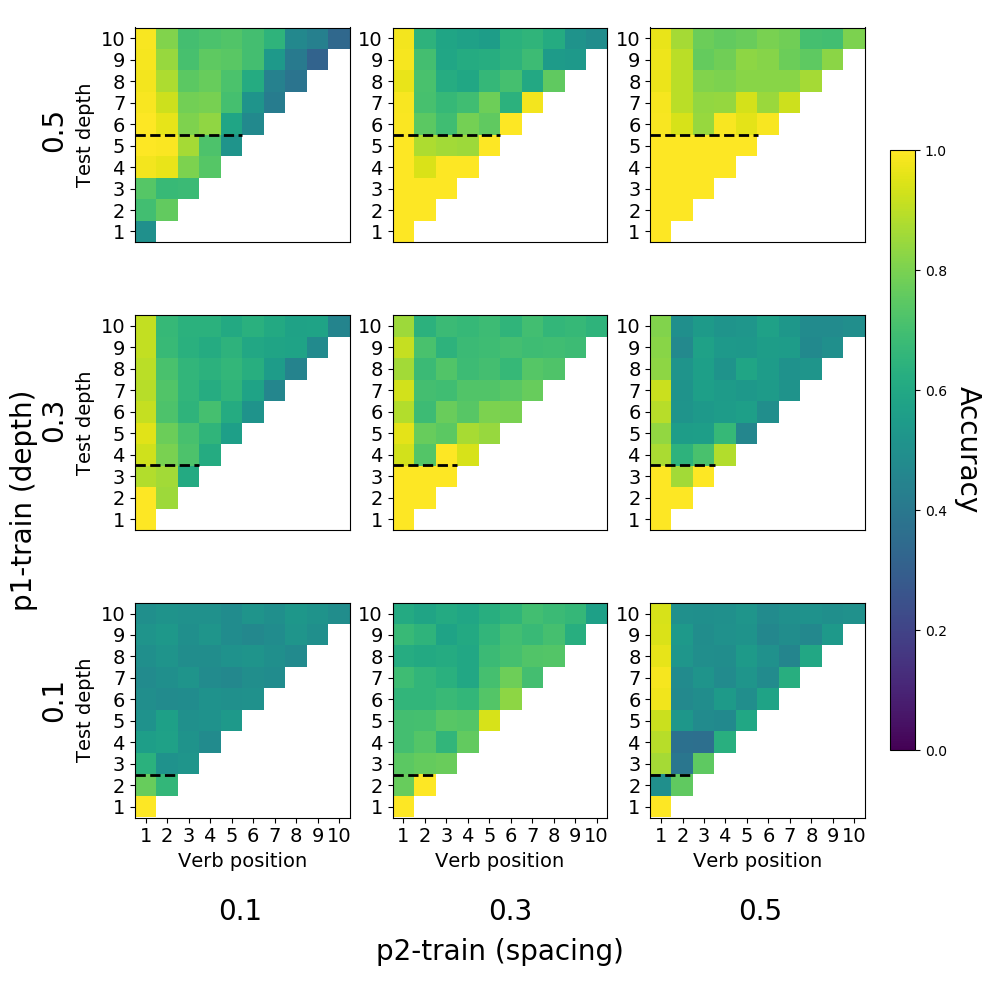}
    \caption{\textbf{Stack-LSTMs: }Accuracy for the nine models on number-agreement tasks with spacing equals to two ($s=2$). Each matrix corresponds to a model trained on one of the $(p1, p2)$-training datasets. Each pixel in a matrix corresponds to accuracy on a specific verb in the sentence when tested on a NA-task with a given depth $d$. Dashed horizontal lines correspond to the maximal depth observed during training.}
    \label{fig:}
\end{figure*}